\documentclass[sigconf]{acmart}

\graphicspath{ {../figs/} }

\AtBeginDocument{%
  \providecommand\BibTeX{{%
    \normalfont B\kern-0.5em{\scshape i\kern-0.25em b}\kern-0.8em\TeX}}}

\copyrightyear{2021}
\acmYear{2021}
\setcopyright{acmlicensed}
\acmConference[KDD '21]{Proceedings of the 27th ACM SIGKDD Conference on Knowledge Discovery and Data Mining}{August 14--18, 2021}{Virtual Event, Singapore}
\acmBooktitle{Proceedings of the 27th ACM SIGKDD Conference on Knowledge Discovery and Data Mining (KDD '21), August 14--18, 2021, Virtual Event, Singapore}
\acmPrice{15.00}
\acmDOI{10.1145/3447548.3467273} 
\acmISBN{978-1-4503-8332-5/21/08}

\usepackage{subfloat}
\usepackage{amsfonts}
\usepackage{amsmath}
\usepackage{booktabs}
\usepackage{multirow}
\usepackage{multicol}
\usepackage{subfigure}
\usepackage{comment}
\usepackage{algpseudocode}
\usepackage{algorithm,algorithmicx}
\usepackage{amsthm}
\usepackage{color}
\usepackage[shortlabels]{enumitem}

\begin{document}

\fancyhead{}

\title{Privacy-Preserving Representation Learning on Graphs: \\ A Mutual Information Perspective}

\author{Binghui Wang}
\affiliation{%
\institution{Illinois Institute of Technology \\ \& Duke University}
  \country{bwang70@iit.edu}
}

\author{Jiayi Guo}
\affiliation{%
  \institution{Tsinghua University}
  \country{guo-jy20@mails.tsinghua.edu.cn}
}

\author{Ang Li, Yiran Chen, and Hai Li}
\affiliation{%
  \institution{Duke University}
  \country{{ang.li630,yiran.chen,hai.li}@duke.edu}
}

\begin{abstract}
    Learning with graphs
    has attracted significant attention recently. 
    Existing representation learning methods on graphs have achieved state-of-the-art performance on various graph-related tasks such as node classification, link prediction, etc. However, we observe that these methods could leak serious private information. 
    For instance, one can accurately infer the links (or node identity) in a graph from a node classifier (or link predictor) trained on the learnt node representations by existing methods.
    To address the issue, we propose a privacy-preserving representation learning framework on graphs from the \emph{mutual information} perspective. Specifically, our framework includes a primary learning task and a privacy protection task, and we consider node classification and link prediction as the two tasks of interest. Our goal is to learn node representations such that they can be used to achieve high performance for the primary learning task, while obtaining  performance for the privacy protection task close to random guessing. 
    We formally formulate our goal via mutual information objectives. However, it is intractable to compute mutual information in practice. Then, we derive tractable variational bounds for the mutual information terms, where each bound can be parameterized via a neural network. Next, we train these parameterized neural networks to approximate the true mutual information and learn privacy-preserving node representations. 
    We finally evaluate our framework on 
    various graph datasets.
\end{abstract}



\maketitle

\section{Introduction}
\label{sec:intro}

Graph is a powerful tool to represent diverse types of data including social networks, chemical networks, etc. 
Learning with graphs has been an active research topic recently and various representation learning methods on graphs~\cite{perozzi2014deepwalk,tang2015line,cao2015grarep,tu2016max,yang2016revisiting,grover2016node2vec,kipf2017semi,duran2017learning,velivckovic2018graph,rossi2018deep,hamilton2017inductive,xu2018powerful,ribeiro2017struc2vec,xu2018representation,qu2019gmnn,ma2019graph,liu2019hyperbolic,chami2019hyperbolic,wu2019simplifying,cui2020adaptive} have been proposed.
Given a graph, representation learning on the graph aims to learn a node embedding function that maps each node in the observational space 
to a latent space by capturing the graph structural information. 
The learned node embedding function can be used for many graph-related tasks.
For instance, node classification and link prediction are two basic tasks. 
Node classification aims to predict the label of unlabeled nodes in a graph based on a set of labeled nodes, while link prediction aims to predict the link status between a pair of nodes, based on a set of observed positive/negative links\footnote{Positive means there exists a link between a pair of nodes, and negative means no link between them.} in a graph. 

Existing representation learning methods on graphs have demonstrated to achieve state-of-the-art performance on these tasks, e.g., node classification~\cite{kipf2017semi,velivckovic2018graph} and link prediction~\cite{zhang2018link}. 
However, due to that the learnt node  representations are not task-specific, we note that existing methods 
could unintendedly leak important information. 
For instance, we observe that one can 
accurately infer the links in a graph from a node classifier trained on the learnt node  representations; and one can also predict the node labels from a link predictor based on the node representations (See Table~\ref{tbl:link_pred_w_node_prot} in Section~\ref{eval:results}). 
Such information leakage could involve serious privacy issues. 
Take users in a social network (e.g., Twitter) as an example. 
Some users (e.g., celebrities) in the social network may just want to make their identities known to the public, but they do not want to expose their private social relationship (e.g., family relationship). 
Some other users (e.g., malicious users) do not want to reveal their identities, but do want to expose their social relationship with normal users to make themselves also look normal. 
Suppose the social network has deployed a user identity classification system (i.e., node classification) or friendship recommendation  system (i.e., link prediction)
using certain graph representation learning method. 
Then, if an adversary (e.g., insider) knows the method, he could thus infer the user's private friendship links/identities.

\noindent {\bf Our work:} In this paper, we aim to address the above privacy violation issue and propose a privacy-preserving representation learning framework on graphs from the \emph{mutual information} perspective. 
Specifically, our framework 
includes a primary learning task and a privacy protection task. 
We consider node classification and link prediction as the two tasks of interest\footnote{Note that our framework can be generalized to any graph-related tasks.}. 
Under this context, our framework includes three modules: node embedding function (for node representation learning), link predictor (uses the node representations to perform link prediction), and node classifier 
(uses node representations to perform node classification). 

Then, we target the following two problems:

\begin{itemize}
\item {\bf Problem 1: Link prediction + node privacy protection.} The primary learning task is learning node representations such that the link predictor can achieve high link prediction performance, and the privacy protection task is to enforce that the learnt node representations cannot be used by the node classifier to accurately infer the node label. 

\item {\bf Problem 2: Node classification + link privacy protection.} The primary learning task is learning node representations such that the node classifier can achieve high node classification performance, and the privacy protection task is to enforce that the learnt node representations cannot be used by the link predictor to accurately infer the link status.  
\end{itemize}

We formally formulate our problems using two \emph{mutual information} objectives, which are defined on the primary learning task and the privacy protection task, respectively. 
Then, for Problem 1, the goal of the two objectives is to 
learn an embedding function such that: 
1) The representations 
of a pair of nodes retain as much information as possible of the respective link status. 
Intuitively, when the pair of node  representations keep the most information about the link status, the link predictor trained on the node representations could have the highest link prediction performance. 
2) The node representation contains as less information as possible about the node label. 
Intuitively, when the learnt node representation preserves the least information on the node label, the node classifier trained on the node representations could have the lowest node classification performance. 
Similarly, for Problem 2, the goal is to learn an embedding function such that: 
1) The node representation contains as much information as possible to facilitate predicting the node label. 
2) The representations of node pairs retain as less information as possible to prevent  inferring the link status.  

However, the mutual information terms are challenging to calculate in practice, as they require to compute an intractable posterior distribution. 
Motivated by mutual information neural estimators~\cite{belghazi2018mutual,chen2016infogan,cheng2020club}, we convert the intractable mutual information terms to be the tractable ones via introducing variational (upper and lower) bounds. Specifically, each variational bound involves a variational posterior distribution, and it be parameterized via a neural network. Estimating the true mutual information thus reduces to training the parameterized neural networks. Furthermore, we propose an alternative training algorithm to train these neural networks.

We finally evaluate our framework on multiple benchmark graph datasets.
Experimental results demonstrate that without privacy protection, the learnt node representations by existing methods for the primary learning task can be also used to obtain high performance on the privacy protection task.  
However, with our proposed privacy-protection mechanism, the learnt node representations can only be used to achieve high performance for the primary learning task, while obtaining the performance for the privacy protection task close to random guessing. 
Our key contributions can be summarized as follows:
\begin{itemize}
    \item We propose the first work to study privacy-preserving representation learning on graphs.
    
    \item We formally formulate our problems via mutual information objectives and design tractable algorithms to estimate intractable mutual information. 
    
    \item We evaluate our framework on various graph datasets and 
    results demonstrate the effectiveness of our framework for privacy-preserving representation learning on graphs.  
\end{itemize}

\section{Related Work}
\label{sec:related}

\subsection{Representation Learning on Graphs}

Various representation learning methods on graphs have been proposed~\cite{perozzi2014deepwalk,tang2015line,cao2015grarep,tu2016max,yang2016revisiting,grover2016node2vec,kipf2017semi,duran2017learning,velivckovic2018graph,rossi2018deep,hamilton2017inductive,xu2018powerful,ribeiro2017struc2vec,xu2018representation,qu2019gmnn,ma2019graph,wu2019net,liu2019hyperbolic,chami2019hyperbolic,wu2019simplifying,cui2020adaptive} in the past several years. 
Graph representation learning based on graph neural networks have exhibit stronger performance than random walk- and factorization-based  methods~\cite{perozzi2014deepwalk,grover2016node2vec,tang2015line,cao2015grarep,qiu2018network}.
For instance, Graph Convolutional Network (GCN)~\cite{kipf2017semi} is motivated by spectral graph convolutions~\cite{duvenaud2015convolutional} and learns node representations, based on the graph convolutional operator, for node classification. 
HGCN~\cite{chami2019hyperbolic}
leverages both the expressiveness of GCN and hyperbolic geometry to learn node representations. 
Specifically, HGCN designs GCN operations in the hyperbolic space and maps Euclidean node features to embeddings in hyperbolic spaces with trainable curvatures at each layer. 
The learnt node representations make HGCN achieve both higher node classification performance and link prediction performance than Euclidean space-based GCNs. 

A few recent works~\cite{velickovic2019deep,sun2020infograph,peng2020graph} propose to leverage mutual information to perform \emph{unsupervised} graph representation learning. 
For instance, 
Peng et al.~\cite{peng2020graph} propose a concept called Graphical Mutual Information (GMI), which measures the correlation between the entire graph and high-level hidden representations, and is invariant to the isomorphic transformation of input graphs. 
By virtue of GMI, the authors design an unsupervised model trained by maximizing GMI between the input and output of a graph neural encoder. The learnt node representations of GMI are used for node classification and link prediction and GMI achieves better performance than other unsupervised graph representation learning methods. 

Note that although our framework also adopts mutual information, its goal is completely different from mutual information-based graph representation learning methods. Our goal is to learn privacy-preserving node representations that consider both a primary learning task and a privacy protection task, while these existing methods mainly focus on learning node representations that achieve high performance for a primary learning task.

\subsection{Mutual Information Estimation}
Estimating mutual information accurately between high dimensional continuous random variable is challenging~\cite{belghazi2018mutual}.
To obtain differentiable and scalable mutual information estimation, recent methods~\cite{alemi2017deep,belghazi2018mutual,oord2018representation,poole2019variational,hjelm2019learning,cheng2020club} propose to first derive mutual information 
(upper or lower) 
bounds by introducing auxiliary variational  distributions and then train parameterized neural networks to estimate variational distributions and approximate true mutual information.  
For instance, MINE~\cite{belghazi2018mutual}
treats mutual information as the KL divergence between the joint and marginal distributions, converts it into the dual representation, and obtains a lower mutual information bound. 
Cheng et al.~\cite{cheng2020club} propose a Contrastive Log-ratio Upper Bound (CLUB) of mutual information. CLUB bridges mutual information estimation with contrastive learning~\cite{oord2018representation}, and  mutual information is estimated by the difference of conditional probabilities between positive and negative sample pairs.

\subsection{Other Privacy-Preserving Techniques} 
Differential privacy (DP) and homomorphic encryption (HE) are two other types of methods that ensure privacy protection. However, DP incurs utility loss and HE incurs intolerable computation overheads.
Mutual information is a recent methodology that protects privacy 
based on information theory~\cite{li2020tiprdc}. Compared to DP and HE, the mutual information-based method is demonstrated to be more efficient or/and effective. Motivated by these advantages, we adopt mutual information to study privacy-preserving graph representation learning.

\section{Background \& Problem Definition}
\label{background}

\subsection{Representation Learning on Graphs}
Let $G=(\mathcal{V}, \mathcal{E}, \mathbf{A}, \mathbf{X})$ be an attributed graph, where $v \in \mathcal{V}$ is a node and $N = |V|$ is the total number of nodes; $ (u, v) \in \mathcal{E}$ is a link between $u$ and $v$; $\mathbf{A} \in \mathbb{R}^{N \times N}$ is the adjacency matrix, where $A_{u,v} = 1$, if $(u,v) \in \mathcal{E}$ and $A_{u,v} = 0$, otherwise; and $\mathbf{X} = [\mathbf{x}_1, \mathbf{x}_2, \cdots, \mathbf{x}_{N}] \in \mathbb{R}^{D \times N} $ is the 
node feature matrix with $\mathbf{x}_u \in \mathbb{R}^D$ the node $u$'s feature vector. 
The purpose of representation learning on graphs is to learn a node embedding function 
$f_\theta: \mathbb{R}^{D} \times \mathbb{R}^{N \times N} \rightarrow \mathbb{R}^{d}$, parameterized by $\theta$, that maps each node $u$'s feature vector $\mathbf{x}_u$ in the observational space to a feature vector $\mathbf{z}_u$ in a latent space by capturing the graph structural information, i.e., $\mathbf{z}_u =  f_\theta (\mathbf{x}_u,\mathbf{A}) \in \mathbb{R}^{d}$, where we call $\mathbf{z}_u$ the \emph{node representation}. The learnt node representations can be used for various graph-related tasks.
In this paper, we mainly focus on  two tasks of interest: node classification and link prediction. 

{\bf Node classification.} Each node $v \in V$ in the graph $G$ is associated with a label $y_v$ from a label set $\mathcal{Y} = \{1, 2, \cdots, C \}$. 
Then, given a set of $\mathcal{V}_L \subset \mathcal{V}$ labeled nodes 
with the node representations 
$\{(\mathbf{z}_u, y_u)\}_{u \in \mathcal{V}_L}$ 
as the training nodes,  
node classification is to take the training nodes and their learnt representations as input
and learn a node classifier $g_\psi: \mathbb{R}^{d} \rightarrow \mathbb{R}^{|\mathcal{Y}|}$, parameterized by $\psi$, that has a minimal loss on the training nodes. 
Suppose we use the cross-entropy loss. Then, the objective function of node classification is defined as follows:
{\small
\begin{align*}
    \min_{\psi} \sum_{v \in \mathcal{V}_L} CE(g_{\psi} (\mathbf{z}_v), y_v) = - \sum_{v \in \mathcal{V}_L} 
    \mathbf{1}_{y_v} \circ  \log g_{\psi} (\mathbf{z}_v),
\end{align*}
}%
where $\mathbf{1}_{y_v}$ is an indicator vector whose $y_v$-th entry is 1, and 0, otherwise. 
With the learnt $\psi^*$, we can predict 
the label for each unlabeled node $u \in \mathcal{V} \setminus \mathcal{V}_L$
as $\hat{y}_u = \arg \max_{i} \, g_{\psi^*}(\mathbf{z}_u)_{i}$.

{\bf Link prediction.}
Given a set of positive links $\mathcal{E}_p \subset \mathcal{E} $  (i.e., $A_{uv}=1$) and a set of negative links $\mathcal{E}_n \not \subset \mathcal{E} $ (i.e., $A_{uv}=0$) as the training links. 
Link prediction is to take the training links and the associated  nodes' representations as input
and learn a link predictor $h_\phi: \mathbb{R}^{d} \times \mathbb{R}^{d} \rightarrow [0,1]$, parameterized by $\phi$, that has a minimal reconstruction error on the training links. 
Specifically, the objective function of link prediction we consider is as follows:
{%
\small
\begin{align*}
    \min_{\phi} \sum_{(u,v) \in \mathcal{E}_p \cup \mathcal{E}_n} CE(h_\phi(\mathbf{z}_u, \mathbf{z}_v), A_{uv}) \notag 
    = \sum_{(u,v) \in \mathcal{E}_p \cup  \mathcal{E}_n} - A_{{uv}} \circ  \log h_\phi(\mathbf{z}_u, \mathbf{z}_v).
\end{align*}
}%
With the learnt $\phi^*$, we  predict a link between unlabeled pair of nodes $u$ and $v$ if $h_\phi^*(\mathbf{z}_u, \mathbf{z}_v) > 0.5$, predict no link, otherwise. 

\subsection{Problem Definition}
Node classification and link prediction are two graph-related tasks. 
However, in existing graph representation learning methods, given a primary learning task, one can also obtain promising performance for the other task with the learnt node representations. 
That is, one can accurately infer the link status between nodes (or infer the node label) even if the primary learning task is node classification (or link prediction) (See Table~\ref{tbl:link_pred_wo_node_prot} and Table~\ref{tbl:node_pred_wo_link_prot} in Section~\ref{eval:results}). 
Such a phenomenon could induce privacy concerns in practical applications.
For instance, a celebrity in Twitter just wants to share his identity, but does not want to reveal his private family relationship. 
A malicious user in Twitter does not want to expose his identity, but does want to make his social relationship with normal users known to the public, in order to let himself also look normal.

We highlight that the root cause of the above consequences is that when learning node representations for a task, existing methods 
do not consider protecting privacy for other tasks. 
To address the issue, we are motivated to propose privacy-preserving representation learning methods on graphs. 
We mainly consider the node classification and link prediction tasks, where one is the primary learning task and the other is the privacy protection task. 
Therefore, our problem involves three modules: node embedding function (for node representation learning), link predictor (uses the node representations to perform link prediction), and node classifier (uses node representations to perform node classification). 
In particular, we study the following two problems, each involving a primary learning task and a privacy protection task. 
\begin{itemize}
    \item {\bf Problem 1: Link prediction + node privacy protection.} In this problem, our primary learning task is learning node representations 
    such that the link predictor can achieve high link prediction performance, and our privacy protection task is to enforce that the learnt node representations cannot be used by the node classifier to accurately infer the node label. 
    \item {\bf Problem 2: Node classification + link privacy protection.} In this problem, our primary learning task is learning node representations such that the node classifier can achieve high node classification performance, and our privacy protection task is to enforce the learnt node representations cannot be used by the link predictor to accurately infer the link status.  
\end{itemize}
In the next section, we will formally formulate our two problems and design algorithms to solve the problems.

\begin{figure*}[t]
  \includegraphics[width=0.75\textwidth]{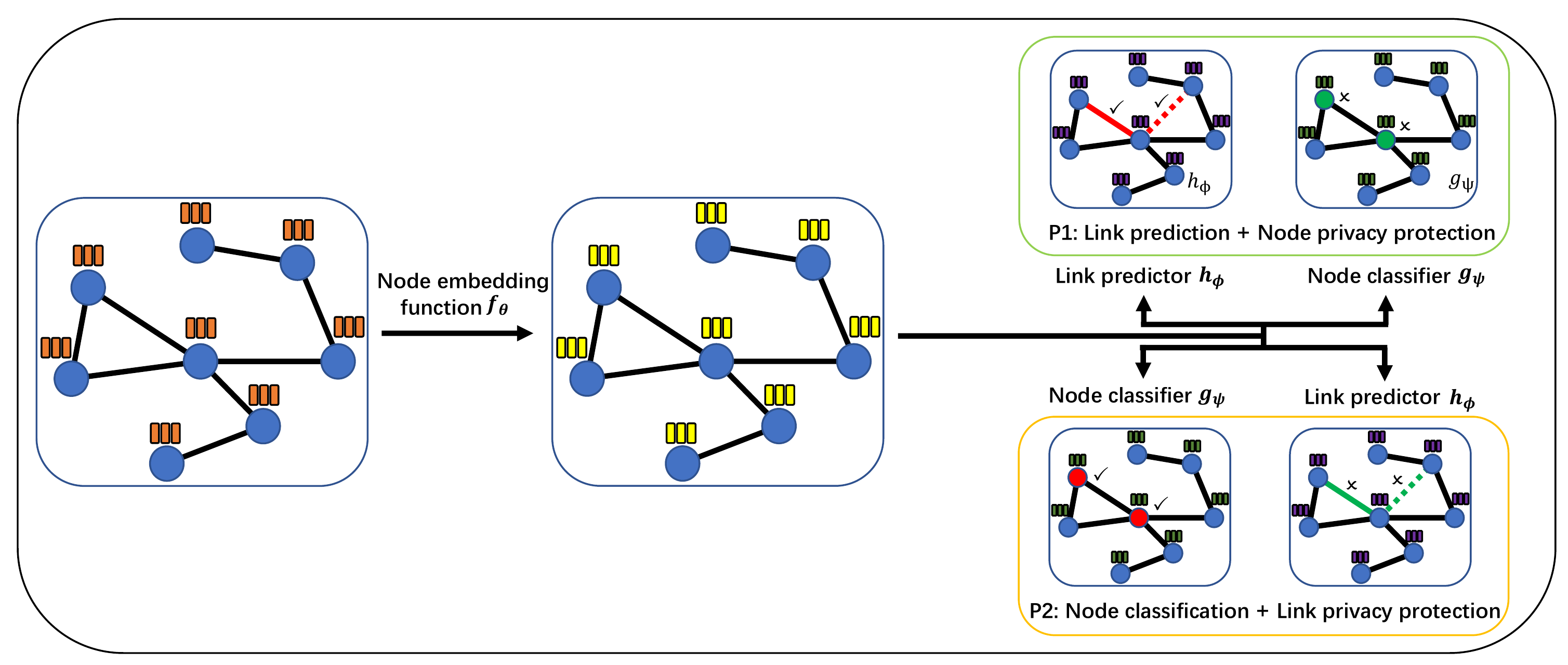}
  \vspace{-3mm}
  \caption{Overview of our privacy-preserving graph representation learning framework. It consists of a primary learning task and a privacy protection task. Our aim is to learn node representations such that they can be used to achieve high performance for the primary learning task, while obtaining performance for the privacy protection task close to random guessing. }
  \label{fig:overview}
\end{figure*}

\section{Privacy-Preserving Representation Learning on Graphs}

We formulate our problems via \emph{mutual information}. Specifically,  
we define two mutual information objectives that are associated with the primary learning task and the privacy protection task, respectively. 
However, the mutual information terms are challenging to calculate in practice. 
Then, we convert them to be the tractable ones via designing variational bounds, and each bound can be estimated by a parameterized neural network. 
Finally, we propose algorithms to train these neural networks to achieve high performance for the primary learning task, and performance close to random guessing for the privacy protection task. 
Figure~\ref{fig:overview} overviews our privacy-preserving graph representation learning framework.

\subsection{Problem 1: Link Prediction with Node Privacy Protection}
\label{sec:node_protect}

\subsubsection{Formulating Problem 1 with mutual information objectives}
Suppose we have a set of samples $\{\mathbf{x}_u$, $y_u$, $A_{uv}\}$ 
consisting of node features $\mathbf{x}_u$, node label $y_u$, link status $A_{uv}$. 
We define the probability distribution associated with a node (i.e., node features $\mathbf{x}_u$ and node label $y_u$) as $p(\mathbf{x}_u, y_u)$. 
Moreover, we define the probability distribution associated with a link (i.e., a pair of node features $\mathbf{x}_u$ and $\mathbf{x}_v$, and the associated link status $A_{uv}$) as $p(\mathbf{x}_u, \mathbf{x}_v, A_{uv})$\footnote{Note that for notation simplicity, we slightly abuse the notations
 $A_{uv}$, $\mathbf{x}_u$, and $y_u$. 
That is, these notations are originally used for the link status between nodes $u$ and $v$, $u$'s raw feature vector and $u$'s label in the graph $G$. Here, we also use them to indicate random variable/vector.}.

Our goal is to learn an embedding function $f_\theta$ to transform 
$\mathbf{x}_u$ to the representation $\mathbf{z}_u = f_\theta(\mathbf{x}_u)$ such that: 
1) The node representation of a pair (e.g., $u$ and $v$) retain as much information as possible on the link status (e.g., $A_{uv}$). 
Intuitively, when 
the representations of the node pair
keep the most information about the link, the link predictor $h_\phi$ trained on the node representations could have the highest link prediction performance.
2) The node representation (e.g., $u$) contains as less information as possible about the node label (e.g., $y_u$).
Intuitively, when the node representation preserves the least information on the node label, the node classifier $g_\psi$ trained on the node representation could have the lowest node classification performance. 
Formally, to achieve 1) and 2), we have the following two respective mutual information objectives: 
{
\begin{align}
    & \textbf{Link prediction: } 
    \max \limits_{\theta} I(A_{uv};\mathbf{z}_u,\mathbf{z}_v), \label{eqn:link_pred} \\
    & \textbf{Node privacy protection: } 
    \min \limits_{\theta}I(\mathbf{z}_u;y_u),  \label{eqn:node_prot}
\end{align}
}%
where $\mathbf{z}_u = f_\theta(\mathbf{x}_u)$ is the random vector after applying the embedding function on  $\mathbf{x}_u$.  
$I(A_{uv};\mathbf{z}_u,\mathbf{z}_v)$ is the mutual information between $A_{uv}$ and the joint ($\mathbf{z}_u$, $\mathbf{z}_v$), which indicates the information  ($\mathbf{z}_u$, $\mathbf{z}_v$) kept for the link variable $A_{uv}$. 
We maximize such mutual information 
to enhance the link prediction performance. 
$I(\mathbf{z}_u;y_u)$ is the mutual information between $\mathbf{z}_u$ and $y_u$ and indicates the information  $\mathbf{z}_u$ preserves for the label variable $y_u$. We minimize such mutual information 
to protect node privacy. Ideally, if $I(\mathbf{z}_u;y_u)=0$, no node's label can be inferred from its node representation, i.e., no node  classifier can perform better than random guessing. 

\subsubsection{Estimating mutual information via tractable variation lower bound and upper bound.}
In practice, the mutual information terms in Equations~\ref{eqn:link_pred} and~\ref{eqn:node_prot} are hard to compute 
as the random variables are potentially high-dimensional and mutual information terms require to know posterior distributions that are challenging to calculate. 
To address the challenge, we are inspired by existing mutual information neural estimation methods~\citep{alemi2017deep,belghazi2018mutual,oord2018representation,poole2019variational,hjelm2019learning,cheng2020club}, which convert the intractable mutual information calculation to the tractable one by designing variational bounds. 
Specifically, we first obtain a mutual information variational lower bound for Equation~\ref{eqn:link_pred} and a mutual information variational upper bound for Equation~\ref{eqn:node_prot} by introducing two auxiliary posterior distributions, respectively.  
Then, we parameterize each auxiliary distribution with a neural network, and approximate the true posteriors by maximizing the variational lower bound and minimizing the variational upper bound through training the involved neural network. 

\noindent {\bf Maximizing the mutual information in Equation~\ref{eqn:link_pred}.}
To solve Equation~\ref{eqn:link_pred}, we derive the following 
variational lower bound:
{
\begin{equation}
    \begin{split}
        & I(A_{uv};\mathbf{z}_u,\mathbf{z}_v) \\
        = & H(A_{uv}) - H(A_{uv}|\mathbf{z}_u, \mathbf{z}_v) \\
        = & H(A_{uv}) + \mathbb{E}_{  p(\mathbf{z}_u, \mathbf{z}_v, A_{uv})}[\log p(A_{uv}|\mathbf{z}_u, \mathbf{z}_v)]]\\
        = & H(A_{uv}) + \mathbb{E}_{   p(\mathbf{z}_u, \mathbf{z}_v,A_{uv})} [KL\big(p(\cdot |\mathbf{z}_u, \mathbf{z}_v)||q_\phi(\cdot |\mathbf{z}_u,\mathbf{z}_v)\big)] \\
        & \qquad \qquad + \mathbb{E}_{ p( \mathbf{z}_u, \mathbf{z}_v,A_{uv})}[\log q_\phi(A_{uv}|\mathbf{z}_u, \mathbf{z}_v)]]\\
        \geq & H(A_{uv}) + \mathbb{E}_{ p(\mathbf{z}_u,  \mathbf{z}_v, A_{uv})}[\log q_\phi(A_{uv}|\mathbf{z}_u, \mathbf{z}_v)]\\
        := & I_{vLB}(A_{uv}; \mathbf{z}_u, \mathbf{z}_v), \label{MI_LB} \\
    \end{split}
\end{equation}
}%
where $KL[q(\cdot)||p(\cdot)]$ is the Kullback-Leibler divergence between two distributions $q(\cdot)$ and $p(\cdot)$ and is nonnegative. 
$q_{\phi}$ is an (arbitrary) auxiliary posterior distribution. $I_{vLB}(A_{uv}; \mathbf{z}_u, \mathbf{z}_v)$ is the variational lower bound of the true mutual information and $H(A_{uv})$ is a constant. 
Note that the lower bound is tight when the auxiliary distribution $q_{\phi}$ becomes the true posterior distribution $p$. 

Our target now is to maximize the lower bound by estimating the auxiliary posterior distribution $q_\phi$ via a parameterized neural network. 
Specifically, we have
{
\begin{align}
        &\max \limits_{\theta} I_{vLB}(A_{uv};\mathbf{z}_u,\mathbf{z}_v) \notag \\
        \Leftrightarrow &  \max_{\theta} \max_{\phi} \mathbb{E}_{    p(A_{uv},\mathbf{z}_u, \mathbf{z}_v)}
        [\log q_\phi({A}_{uv}|\mathbf{z}_u,\mathbf{z}_v)] \notag \\
        = & \max_{\theta} \max_{\phi} \mathbb{E}_{    p(A_{uv}, \mathbf{x}_u, \mathbf{x}_v)}
        [\log q_\phi({A}_{uv}|f_\theta(\mathbf{x}_u), f_\theta( \mathbf{x}_v))] \label{eqn:link_pred_var} 
\end{align}
}%

\noindent {\bf Minimizing the mutual information in Equation~\ref{eqn:node_prot}.}
To solve Equation~\ref{eqn:node_prot}, we leverage the variational upper bound   in~\cite{cheng2020club}:
\begin{equation}
    \begin{split}
        &I(\mathbf{z}_u;y_u) \\
        \leq & I_{vCLUB}(\mathbf{z}_u;y_u) \notag \\ 
        = & \mathbb{E}_{p(\mathbf{z}_u, y_u)} [\log q_{\psi}(y_u|\mathbf{z}_u) ] - \mathbb{E}_{ p(\mathbf{z}_u) p(y_u)} [\log q_{\psi}(y_u|\mathbf{z}_u) ], 
    \end{split}
\end{equation}
where 
$q_{\psi}(y_u | \mathbf{z}_u)$ is an auxiliary distribution of $p(y_u | \mathbf{z}_u)$ that needs to satisfy the following condition~\citep{cheng2020club}: 
\begin{align}
\label{eqn:KL_cond}
KL (p(\mathbf{z}_u, y_u) ||  q_{\psi} (\mathbf{z}_u, y_u)) \leq KL (p(\mathbf{z}_u)p(y_u) || q_{\psi} (\mathbf{z}_u, y_u)). 
\end{align}
To achieve Inequality~\ref{eqn:KL_cond}, 
we need to 
minimize:
{
\begin{align}
    & \min_{\psi}  KL (p(\mathbf{z}_u, y_u) ||  q_{\psi} (\mathbf{z}_u, y_u)) \notag \\
    =& \min_{\psi}  KL (p(y_u | \mathbf{z}_u) ||  q_{\psi} (y_u | \mathbf{z}_u)) \notag \\
    =& \min_{\psi}  \mathbb{E}_{    p(\mathbf{z}_u, y_u)} [\log p(y_u| \mathbf{z}_u)] -  \mathbb{E}_{    p(\mathbf{z}_u, y_u)} [\log q_\psi(y_u| \mathbf{z}_u))] \notag \\
    \Leftrightarrow & \max_{\psi} \mathbb{E}_{    p(\mathbf{z}_u, y_u)} [\log q_\psi(y_u| \mathbf{z}_u)], \label{eqn:KL_conv} 
\end{align}
}%
where we have the last Equation because the first term in the second-to-last Equation is irrelevant to $\psi$.   

Finally, achieving Equation~\ref{eqn:node_prot} becomes solving the following adversarial training objective:
\begin{align}
        &\min \limits_{\theta} \min \limits_{\psi}  I_{vCLUB}(\mathbf{z}_u;y_u) \notag \\ 
        \Leftrightarrow & \min \limits_{\theta} \max \limits_{\psi} 
        \mathbb{E}_{    p(\mathbf{z}_u, y_u)} [\log q_\psi(y_u| \mathbf{z}_u)] \notag \\
        = & \min \limits_{\theta} \max \limits_{\psi} 
        \mathbb{E}_{p(\mathbf{x}_u, y_u)} [\log q_\psi(y_u| f_\theta(\mathbf{x}_u))] \label{eqn:node_prot_adv}
\end{align}
\emph{Remark.} The above objective function can be interpreted as an adversarial game between an adversary $q_\psi$ who aims to infer the label $y_u$ from
$\mathbf{z}_u$ and a defender (i.e., the embedding function $f_\theta$) who aims to protect the node privacy from being inferred. 

\noindent {\bf Implementation via training parameterized neural networks.} 
We solve Equation~\ref{eqn:link_pred_var} and Equation~\ref{eqn:node_prot_adv} in practice
via training two parameterized neural networks associated with the two auxiliary posterior distributions $q_\phi$ and $q_\psi$. 
With it, we expect to obtain high link prediction performance for our primary learning task and low node classification performance for our privacy protection task. 

To solve  Equation~\ref{eqn:link_pred_var}, we first sample 
a set of triplets $\{A_{uv}, \mathbf{x}_u, \mathbf{x}_v \}$ from the graph $G$. 
Then, we parameterize the variational posterior distribution 
$q_\phi$ via a link predictor $h_\phi$ defined on the node representations $f_\theta(\mathbf{x}_u)$ and $f_\theta(\mathbf{x}_v)$ of the sampled node pairs $u$ and $v$.
Suppose we are given a set of positive links $\mathcal{E}_p$ and a set of negative links $\mathcal{E}_n$, then we have
{
\begin{align}
    & \max_\theta \max_\phi \mathbb{E}_{p(A_{uv},\mathbf{x}_u, \mathbf{x}_v)} [\log q_\phi (A_{uv} | f_\theta(\mathbf{x}_u), f_\theta(\mathbf{x}_v))] \notag \\
    \approx & \max_\theta \max_\phi  \sum_{(u,v) \in \mathcal{E}_p \cup \mathcal{E}_n} - CE(h_\phi(f_\theta(\mathbf{x}_u), f_\theta(\mathbf{x}_v)), A_{uv}) \notag \\
    = & \min_\theta \min_\phi  \sum_{(u,v) \in \mathcal{E}_p \cup \mathcal{E}_n} CE(h_\phi(f_\theta(\mathbf{x}_u), f_\theta(\mathbf{x}_v)), A_{uv}). \label{eqn:empi_link_pred}
\end{align}
}%
To solve Equation~\ref{eqn:node_prot_adv}, we first sample a set of labeled nodes $\{\mathbf{x}_u, y_u \}$, and then we  parameterize 
$q_\psi$ via a node classifier $g_\psi$ defined on the node representations $\{f_\theta(\mathbf{x}_u)\}$ of these labeled nodes. 
Suppose we sample a set of labeled nodes  $\mathcal{V}_L$, then we have 
{
\begin{align}
    & \min_\theta \max_\psi \mathbb{E}_{p(\mathbf{x}_u, y_u)} [\log q_\psi(y_u| f_\theta(\mathbf{x}_u))]  \notag \\
    \approx & \min_\theta \max_\psi  \sum_{v \in \mathcal{V}_L} -CE(g_\psi(f_\theta(\mathbf{x}_v)), y_v). \label{eqn:empi_node_prot} 
\end{align}
}%
Combining Equation~\ref{eqn:empi_link_pred} and Equation~\ref{eqn:empi_node_prot}, we have the final objective function for our {\bf Problem 1} as follows:
{\small
\begin{align}
    & \min_\theta \big( \lambda \min_\phi  \sum_{(u,v) \in \mathcal{E}_p \cup \mathcal{E}_n} CE(h_\phi(f_\theta(\mathbf{x}_u), f_\theta(\mathbf{x}_v)), A_{uv}) \notag \\
    & \qquad \quad - (1-\lambda) \max_\psi  \sum_{v \in \mathcal{V}_L} CE(g_\psi(f_\theta(\mathbf{x}_v)), y_v) \big), \label{eqn:obj_prob1_final}
\end{align}
}%
where $\lambda$ is a trade-off factor to balance between achieving high link prediction performance and low node classification performance. 

Note that Equation~\ref{eqn:obj_prob1_final} involves optimizing three neural networks: the node embedding function $f_\theta$, the link predictor $h_\phi$, and the node classifier $g_\psi$. We alternatively train the three neural networks. 
Specifically, in each round, we perform several iterations of gradient descent to update $\phi$,
several iterations of gradient ascent to update $\psi$, and several iterations of gradient descent to update $\theta$. 
We iteratively perform these steps until reaching a predefined maximal number of rounds or the convergence condition. 
Algorithm~\ref{alg:P1} 
in Appendix illustrates the  training procedure of these networks.

\subsection{Problem 2: Node Classification with Link Privacy Protection}
\label{sec:link_protect}

\subsubsection{Formulating Problem 2 with mutual Information.}
In this problem, our goal is to learn an embedding function $f_\theta$ to transform $\mathbf{x}_u$ to the representation $\mathbf{z}_u = f_\theta(\mathbf{x}_u)$ such that: 
1) The representation of a node (e.g., $u$) contains as much information as possible to facilitate predicting the node label (e.g., $y_u$). 
2) The representation of node pairs (e.g., $u$ and $v$) retain as less information as possible to prevent inferring the link status (e.g., $A_{uv}$). 
Formally, to achieve 1) and 2), we have the following two mutual information objectives: 
\begin{align}
    & \textbf{Node classification: } 
    \max \limits_{\theta}I(\mathbf{z}_u;y_u), \label{eqn:node_pred}\\
    &\textbf{Link privacy protection: } 
    \min \limits_{\theta} I(A_{uv};\mathbf{z}_u,\mathbf{z}_v) \label{eqn:link_prot},
\end{align}

\subsubsection{Estimating mutual information via tractable variation lower bound and upper bound.}
Similarly, we first obtain a lower bound for Equation~\ref{eqn:node_pred} and an upper bound for Equation~\ref{eqn:link_prot} by introducing two auxiliary posterior distributions, respectively.  
Then, we parameterize each auxiliary distribution with a neural network, and train each neural network to maximize the lower bound or minimize the upper bound, respectively.

\noindent {\bf Maximizing the mutual information in Equation~\ref{eqn:node_pred}.}
To solve Equation~\ref{eqn:node_pred}, we have the following variational lower bound 
{\small
\begin{align}
        & I(\mathbf{z}_u;y_u) \notag \\
        =& H(y_u)-H(y_u|\mathbf{z}_u) \notag \\
        =& H(y_u) + \mathbb{E}_{    p(\mathbf{z}_u, y_u)}[\log P(y_u|\mathbf{z}_u)] \notag \\
        =& H(y_u) + \mathbb{E}_{    p(\mathbf{z}_u, y_u)}[KL(p(\cdot|\mathbf{z}_u) || q_\psi(\cdot|\mathbf{z}_u)] \notag \\ 
        & \qquad + \mathbb{E}_{    p(\mathbf{z}_u, y_u)} \log p(y_u|\mathbf{z}_u)] \notag \\
        \geq & H(y_u) + \mathbb{E}_{    p(\mathbf{z}_u, y_u)}[\log q_\psi(y_u|\mathbf{z}_u)] \notag \\
        := & I_{vLB}(\mathbf{z}_u; y_u).
\end{align}
}%
Note that the variational lower bound is tight when the auxiliary distribution $q_{\psi}$ becomes the true posterior distribution $p$. 
Now, we maximize the variational lower bound to achieve Equation~\ref{eqn:node_pred} by estimating 
$q_\psi$. 
Specifically, we have
{
\begin{align}
    & \max_{\theta} I_{vLB}(\mathbf{z}_u; y_u) \notag \\
    \Leftrightarrow & \max_{\theta} \max_{\psi} \mathbb{E}_{p(\mathbf{z}_u, y_u)}[\log q_\psi(y_u|\mathbf{z}_u)] \notag \\
    = & \max_{\theta} \max_{\psi} \mathbb{E}_{ p(\mathbf{x}_u, y_u)}[\log q_\psi(y_u|f_\theta(\mathbf{x}_u))]. \label{eqn:node_pred_var}
\end{align}
}%

\noindent {\bf Minimizing the mutual information in Equation~\ref{eqn:link_prot}.}
To solve Equation~\ref{eqn:link_prot}, we derive the vCLUB motivated by ~\citep{cheng2020club} and have 
{
\begin{align}
    & I(A_{uv}; \mathbf{z}_u,\mathbf{z}_v) \notag \\
    \leq & I_{vCLUB}(A_{uv}; \mathbf{z}_u,\mathbf{z}_v) \notag \\
 = & \mathbb{E}_{p(A_{uv}, \mathbf{z}_u,\mathbf{z}_v)} [\log q_\phi(A_{uv} | \mathbf{z}_u,\mathbf{z}_v)] \notag \\ & \quad - \mathbb{E}_{ p(\mathbf{z}_u, \mathbf{z}_v) p(A_{uv})} [\log q_\phi(A_{uv} | \mathbf{z}_u,\mathbf{z}_v)], 
\end{align}
}%
where $q_\phi(A_{uv} | \mathbf{z}_u,\mathbf{z}_v)$ is an auxiliary distribution of $p(A_{uv} | \mathbf{z}_u,\mathbf{z}_v)$ that needs to satisfy the following condition:
\begin{align}
    & KL(p(\mathbf{z}_u,\mathbf{z}_v,A_{uv})|| q_\phi(\mathbf{z}_u,\mathbf{z}_v,A_{uv})) \notag \\
    \leq &
    KL(p(\mathbf{z}_u, \mathbf{z}_v) p(A_{uv})|| q_\phi(\mathbf{z}_u,\mathbf{z}_v,A_{uv})). 
    \label{eqn:KL_cond2}
\end{align}
That is, $I_{vCLUB}$ is a mutual information upper bound if the variational joint distribution $q_\phi(\mathbf{z}_u,\mathbf{z}_v,A_{uv})$ is closer to the joint distribution $p(\mathbf{z}_u,\mathbf{z}_v,A_{uv})$ than to $p(\mathbf{z}_u,\mathbf{z}_v) p(A_{uv})$. 

To achieve Inequality~\ref{eqn:KL_cond2}, we need to minimize the KL-divergence 
$KL(p(\mathbf{z}_u,\mathbf{z}_v,A_{uv})|| q_\phi(\mathbf{z}_u,\mathbf{z}_v,A_{uv}))$ 
as follows:
{\small
\begin{align}
    & \min_{\phi} KL(p(A_{uv},\mathbf{z}_u, \mathbf{z}_v) || q_\phi(A_{uv},\mathbf{z}_u, \mathbf{z}_v)) \notag \\
    = & \min_{\phi} KL(p(A_{uv} | \mathbf{z}_u, \mathbf{z}_v) || q_\phi(A_{uv} | \mathbf{z}_u, \mathbf{z}_v)) \notag \\
    = & \min_{\phi} \mathbb{E}_{    p(A_{uv},\mathbf{z}_u, \mathbf{z}_v)} [\log p(A_{uv} | \mathbf{z}_u, \mathbf{z}_v)] 
    -
    \mathbb{E}_{    p(A_{uv},\mathbf{z}_u, \mathbf{z}_v)} [\log q_\phi (A_{uv} | \mathbf{z}_u, \mathbf{z}_v)] \notag \\
    \Leftrightarrow & \max_{\phi} \mathbb{E}_{  p(A_{uv},\mathbf{z}_u, \mathbf{z}_v)} [\log q_\phi (A_{uv} | \mathbf{z}_u, \mathbf{z}_v)]. \notag
\end{align}
}%
Finally, our target to achieve Equation~\ref{eqn:link_prot} becomes the following adversarial training objective:
{
\begin{align}
    & \min_{\theta} \min_{\phi} I_{vCLUB}(A_{uv}; \mathbf{z}_u, \mathbf{z}_v) \notag \\
    \Leftrightarrow & \min_\theta \max_\phi \mathbb{E}_{p(A_{uv},\mathbf{z}_u, \mathbf{z}_u)} [\log q_\phi (A_{uv} | \mathbf{z}_u, \mathbf{z}_v)] \notag \\
    = & \min_\theta \max_\phi \mathbb{E}_{p(A_{uv},\mathbf{x}_u, \mathbf{x}_v)} [\log q_\phi (A_{uv} | f_\theta(\mathbf{x}_u), f_\theta(\mathbf{x}_v))] \label{eqn:link_prot_adv}
\end{align}
}%
\emph{Remark.} The above objective function can be interpreted as an adversarial game between an adversary $q_\phi$ who aims to infer the link $A_{uv}$ from the pair of  node representations
$f_\theta(\mathbf{x}_u)$ and $f_\theta(\mathbf{x}_v)$; 
and a defender (i.e.,
the embedding function $f_\theta$) who aims to protect the link privacy from being inferred.

\noindent {\bf Implementation via training parameterized neural networks.} 
We solve Equation~\ref{eqn:node_pred_var} and Equation~\ref{eqn:link_prot_adv} in practice via training two parameterized neural networks.
With it, we expect to obtain a high node classification performance for our primary learning task and a low link prediction performance for our privacy protection task. 

Similar to solving {\bf  Problem 1},
to solve Equation~\ref{eqn:node_pred_var}, we first sample a set of labeled nodes $\{\mathbf{x}_u, y_u \}$, and then we  parameterize the variational posterior distribution $q_\psi$ via a node classifier $g_\psi$ defined on the node representation $\{f_\theta(\mathbf{x}_u)\}$ of these labeled nodes. 
Suppose we sample a set of labeled nodes  $\mathcal{V}_L$, then we have 
{
\begin{align}
    & \max_\theta \max_\psi \mathbb{E}_{p(\mathbf{x}_u, y_u)} [\log q_\psi(y_u| f_\theta(\mathbf{x}_u))]  \notag \\
    \approx & \max_\theta \max_\psi  \sum_{v \in \mathcal{V}_L} -CE(g_\psi(f_\theta(\mathbf{x}_v)), y_v) \notag \\
    = & \min_\theta \min_\psi  \sum_{v \in \mathcal{V}_L} CE(g_\psi(f_\theta(\mathbf{x}_v)), y_v)
    \label{eqn:empi_node_pred} 
\end{align}
}%
To solve  Equation~\ref{eqn:link_prot_adv}, we first sample 
a set of triplets $\{A_{uv}, \mathbf{x}_u, \mathbf{x}_v \}$ from the graph $G$. 
Then, we parameterize $q_\phi$ via a link predictor $h_\phi$ defined on the node representation $f_\theta(\mathbf{x}_u)$ and $f_\theta(\mathbf{x}_v)$ of the sampled node pairs $u$ and $v$.
Depending on the real scenarios, we can protect a set of positive links with $A_{uv}=1$ or/and a set of negative links with $A_{uv}=0$. In our experiments, we consider protecting both positive links and negative links.  
Suppose we are given a set of positive links $\mathcal{E}_p$ and a set of negative links $\mathcal{E}_n$, then we have
{
\begin{align}
    & \min_\theta \max_\phi \mathbb{E}_{p(A_{uv},\mathbf{x}_u, \mathbf{x}_v)} [\log q_\phi (A_{uv} | f_\theta(\mathbf{x}_u), f_\theta(\mathbf{x}_v))] \notag \\
    \approx & \min_\theta \max_\phi  \sum_{(u,v) \in \mathcal{E}_p \cup \mathcal{E}_n} - CE(h_\phi(f_\theta(\mathbf{x}_u), f_\theta(\mathbf{x}_v)), A_{uv}) \label{eqn:empi_link_prot}
\end{align}
}%
Combining Equation~\ref{eqn:empi_node_pred} and Equation~\ref{eqn:empi_link_prot}, we have the final objective function for our {\bf Problem 2} as follows:
{
\begin{align}
    & \min_\theta \big( \lambda \min_\psi  \sum_{v \in \mathcal{V}_L} CE(g_\psi(f_\theta(\mathbf{x}_v)), y_v)  \notag \\
    & \quad - (1-\lambda)  \max_\phi  \sum_{(u,v) \in \mathcal{E}_p \cup \mathcal{E}_n} CE(h_\phi(f_\theta(\mathbf{x}_u), f_\theta(\mathbf{x}_v)), A_{uv}) \big), 
    \label{eqn:obj_prob2_final}
\end{align}
}%
where $\lambda$ is a trade-off factor to balance between achieving high node classification performance and low link prediction performance. 

Similar to {\bf Problem 1}, our objective function for Problem 2 in Equation~\ref{eqn:obj_prob2_final} involves optimizing three neural networks: the node embedding function $f_\theta$, the link predictor $h_\phi$, and the node classifier $g_\psi$. We alternatively train the three neural networks. Algorithm~\ref{alg:P2} in Appendix 
illustrates the  training procedure of these networks.

\begin{table}[]\renewcommand{\arraystretch}{1.4}
\caption{Dataset statistics.}
\centering
\begin{tabular}{|c|c|c|c|c|c|c|} \hline 
{\bf Dataset}  & {\bf \#Nodes} & {\bf \#Edges} & {\bf \#Features} & {\bf \#Labels} \\ \hline
{\bf Cora}  &  {\bf 2,708} & {\bf 5,429} & {\bf 1,433}   & {\bf 7} \\ \hline
{\bf Citeseer}  &  {\bf 3,327} & {\bf 4,732} & {\bf 3,703}   & {\bf 6} \\ \hline
{\bf Pubmed}  &  {\bf 19,717} & {\bf 44,338} & {\bf 500}   & {\bf 3}  \\ \hline
\end{tabular}
\label{dataset_stat}
\end{table}

\begin{table*}[t]\renewcommand{\arraystretch}{1.1}
\centering
\caption{Results on primary learning task: link prediction +  privacy-protection task: node classification.}
\vspace{-2mm}
 \subfigure[][{\bf Without node privacy protection}]{
\begin{tabular}{|c|c|c|c|c|}
\hline
\multirow{2}{*}{\bf Dataset} & \multirow{2}{*}{\bf Method} & \multicolumn{1}{c|}{\bf Link prediction} & \multicolumn{2}{c|}{\bf Node classification} \\ \cline{3-5} 
    &     &  {\bf AUC}   &   {\bf Acc}        &  {\bf Rand}         \\ \hline
\multirow{3}{*}{\bf Cora} &  {\bf GCN} &89.33\%   &72.00\%  &14.29\%  \\ \cline{2-5} 
 &   {\bf GAT} &92.95\%   &71.80\%  &14.29\%   \\ \cline{2-5} 
 &   {\bf HGCN} &93.78\%   &75.30\%  &14.29\%       \\ \hline
\multirow{3}{*}{\bf Citeseer} &  {\bf GCN} &91.52\%  &67.40\% &16.67\%  \\ \cline{2-5} 
& {\bf GAT} &95.03\%  &67.00\% &16.67\%       \\ \cline{2-5} 
                  &     {\bf HGCN} &94.65\%  &67.20\% &16.67\%    \\ \hline
\multirow{3}{*}{\bf Pubmed} &                   {\bf GCN} &91.43\%   &72.70\%  &33.33\%       \\ \cline{2-5} 
                  &                   {\bf GAT}&94.44\%   &78.50\%  &33.33\% \\ \cline{2-5} 
                  &                   {\bf HGCN}&95.18\%   &76.60\%  &33.33\%   \\ \hline
\end{tabular}
\label{tbl:link_pred_wo_node_prot}
}
\,
\subfigure[][{\bf With node privacy protection using our framework}]{
\begin{tabular}{|c|c|c|c|c|}
\hline
\multirow{2}{*}{\bf Dataset} & \multirow{2}{*}{\bf Method} & \multicolumn{1}{c|}{\bf Link prediction} & \multicolumn{2}{c|}{\bf Node classification} \\ \cline{3-5} 
           &     &  {\bf AUC}  &   {\bf Acc}        &  {\bf Rand}         \\ \hline
\multirow{3}{*}{\bf Cora} &  {\bf GCN} &79.41\%   &28.50\%  &14.29\%  \\ \cline{2-5} 
&   {\bf GAT} &  84.12\%   &21.40\%  &14.29\%   \\ \cline{2-5} 
&   {\bf HGCN} &85.01\%   &14.40\%  &14.29\%    \\ \hline
\multirow{3}{*}{\bf Citeseer} &  {\bf GCN} &85.55\%  &15.40\% &16.67\%    \\ \cline{2-5} 
& {\bf GAT} &85.44\%  &21.40\% &16.67\%          \\ \cline{2-5} 
& {\bf HGCN} &86.51\%  &18.20\% &16.67\%         \\ \hline
\multirow{3}{*}{\bf Pubmed} &   {\bf GCN} &81.24\%   &42.50\%  &33.33\%                \\ \cline{2-5} 
 &    {\bf GAT} &84.65\%   &41.80\%  &33.33\%          \\ \cline{2-5} 
& {\bf HGCN} &85.39\%   &40.70\%  &33.33\%       \\ \hline
\end{tabular}
\label{tbl:link_pred_w_node_prot}}
\end{table*}

\begin{table*}[!t]\renewcommand{\arraystretch}{1.1}
\centering
 \caption{Results on primary learning task: node classification + privacy-protection task: link prediction.}
\vspace{-2mm}
\subfigure[][{\bf Without link privacy protection}]{
\begin{tabular}{|c|c|c|c|c|}
\hline
\multirow{2}{*}{\bf Dataset} & \multirow{2}{*}{\bf Method} & \multicolumn{2}{c|}{\bf Link prediction} & \multicolumn{1}{c|}{\bf Node classification} \\ \cline{3-5} 
  &     &  {\bf AUC}    &{\bf Rand}     &   {\bf Acc}     \\ \hline
\multirow{3}{*}{\bf Cora} &  {\bf GCN} &   82.73\%  &50.00\%  &81.60\%  \\ \cline{2-5} 
            &      {\bf GAT} &77.32\%    &50.00\%  &81.80\%          \\ \cline{2-5} 
            & {\bf HGCN} & 80.83\%  &50.00\%  &79.50\%       \\ \hline
\multirow{3}{*}{\bf Citeseer} & {\bf GCN} &83.30\%   &50.00\%  &67.50\%           \\ \cline{2-5} 
 & {\bf GAT} &82.12\%  &50.00\%  &71.00\%          \\ \cline{2-5} 
& {\bf HGCN} &79.12\%  &50.00\%  &68.50\%          \\ \hline
\multirow{3}{*}{\bf Pubmed} &  {\bf GCN} &78.90\%  &50.00\%  &78.80\%           \\ \cline{2-5} 
  & {\bf GAT}&78.45\%  &50.00\%  &79.20\%        \\ \cline{2-5} 
&  {\bf HGCN} &77.10\%  &50.00\%  &80.00\% \\ \hline
\end{tabular}
\label{tbl:node_pred_wo_link_prot}
}
\, 
\subfigure[][{\bf With link privacy protection using our framework}]{
\begin{tabular}{|c|c|c|c|c|}
\hline
\multirow{2}{*}{\bf Dataset} & \multirow{2}{*}{\bf Method} & \multicolumn{2}{c|}{\bf Link prediction} & \multicolumn{1}{c|}{\bf Node classification} \\ \cline{3-5} 
           &     &  {\bf AUC}        &  {\bf Rand}    &   {\bf Acc}                 \\ \hline
\multirow{3}{*}{\bf Cora} &  {\bf GCN} &49.91\%  &50.00\%  &79.70\%  \\ \cline{2-5}  
 &       {\bf GAT} &50.00\%  &50.00\%  &81.30\% \\ \cline{2-5} 
& {\bf HGCN} &54.49\%   &50.00\%  &75.50\%   \\ \hline
\multirow{3}{*}{\bf Citeseer} &      {\bf GCN}&53.29\%  &50.00\%  &65.80\%  \\ \cline{2-5} 
&     {\bf GAT} &50.00\% &50.00\%  &70.70\% \\ \cline{2-5} 
 &   {\bf HGCN} &49.36\%  &50.00\%  &64.50\%  \\ \hline
\multirow{3}{*}{\bf Pubmed} &   {\bf GCN} &49.57\%  &50.00\%  &78.60\%    \\ \cline{2-5} 
&     {\bf GAT} &50.00\%  &50.00\%  &78.50\% \\ \cline{2-5} 
&    {\bf HGCN} &53.22\% &50.00\%  &78.50\% \\ \hline
\end{tabular}
\label{tbl:node_pred_w_link_prot}
}
\end{table*}

\section{Evaluation}
\label{sec:eval}

\subsection{Experimental Setup}
\label{eval:setup}

\noindent {\bf Dataset description.}
We use three benchmark citation graphs  (i.e., Cora, Citeseer, and Pubmed)~\cite{sen2008collective} to evaluate our method. 
In these graphs, each node represents a documents and each edge indicates a citation between two documents.  
Each document treats the bag-of-words feature as the node feature vector and also has a label. 
Table~\ref{dataset_stat} shows basic statistics of these citation graphs. 

\noindent {\bf Representation learning methods.} We select three graph neural networks, i.e., GCN~\cite{kipf2017semi}, GAT~\cite{velivckovic2018graph}, HGCN~\cite{chami2019hyperbolic} as the representative graph representation learning methods. Each method learns node representations for both node classification and link prediction. 
Specifically, in these methods, the input layer to the second-to-last layer are used for learning node representations. 
Suppose node $u$'s representation is $\mathbf{z}_u$.  
When performing node classification, all these methods train a (Euclidean) softmax classifier $g_\psi$ in the last layer, i.e.,
$g_{\psi}(\mathbf{z}_u)= \textrm{softmax}(\mathbf{z}_u^T \cdot \psi)$, where $T$ is a transpose and $\textrm{softmax}(\mathbf{x})_i = \frac{\exp(x_i)}{\sum_{j} \exp(x_j)}$. 
When performing link prediction, GCN and GAT train a parameterized bilinear link predictor $h_\phi$, i.e., $h_\phi(\mathbf{z}_u, \mathbf{z}_v) = \mathbf{z}_u^T \cdot \phi \cdot \mathbf{z}_v$; 
HGCN trains a Fermi-Dirac  decoder~\cite{krioukov2010hyperbolic,nickel2017poincare} as the link predictor. 

Note that these methods only focus on learning node representations for solving the primary learning task and do not consider protecting the privacy for the other task. 
Furthermore, we apply our framework to learn privacy-preserving node representations.

\noindent{\bf Training set, validation set, and testing set.} 
Following existing works~\cite{chami2019hyperbolic,zhang2018link,kipf2017semi}, in each graph dataset, for node classification, we randomly sample 20 nodes per class to form the training set, randomly sample 500 nodes in total as the validation set, and randomly sample 1,000 nodes in total as the testing set. 
For link prediction, we randomly sample 85\% positive links and 50\% negative links for training, sample 5\% positive links and an equal number of negative links for validation, and use the remaining 10\% positive links and sample an equal number of negative links for testing. 

\noindent {\bf Parameter setting.}
We train our framework on the training set and tune hyperparameters to select the model with the minimal error on the validation set. Then, we use the selected model to evaluate the testing set.
By default, we set the trade-off factor $\lambda$ to be 0.5 during training. We also study the impact of $\lambda$ in our experiments. 
We train all graph neural networks using the publicly available source code.
We implement our framework in PyTorch.

\noindent {\bf Evaluation metric.}
Following previous works~\cite{chami2019hyperbolic,zhang2018link,kipf2017semi,kipf2016variational}, we use \emph{AUC} to evaluation the link prediction performance and use \emph{accuracy} to evaluate the node classification performance.

\begin{figure*}[t]
\center
\subfigure[Cora]{\includegraphics[width=0.28\textwidth]{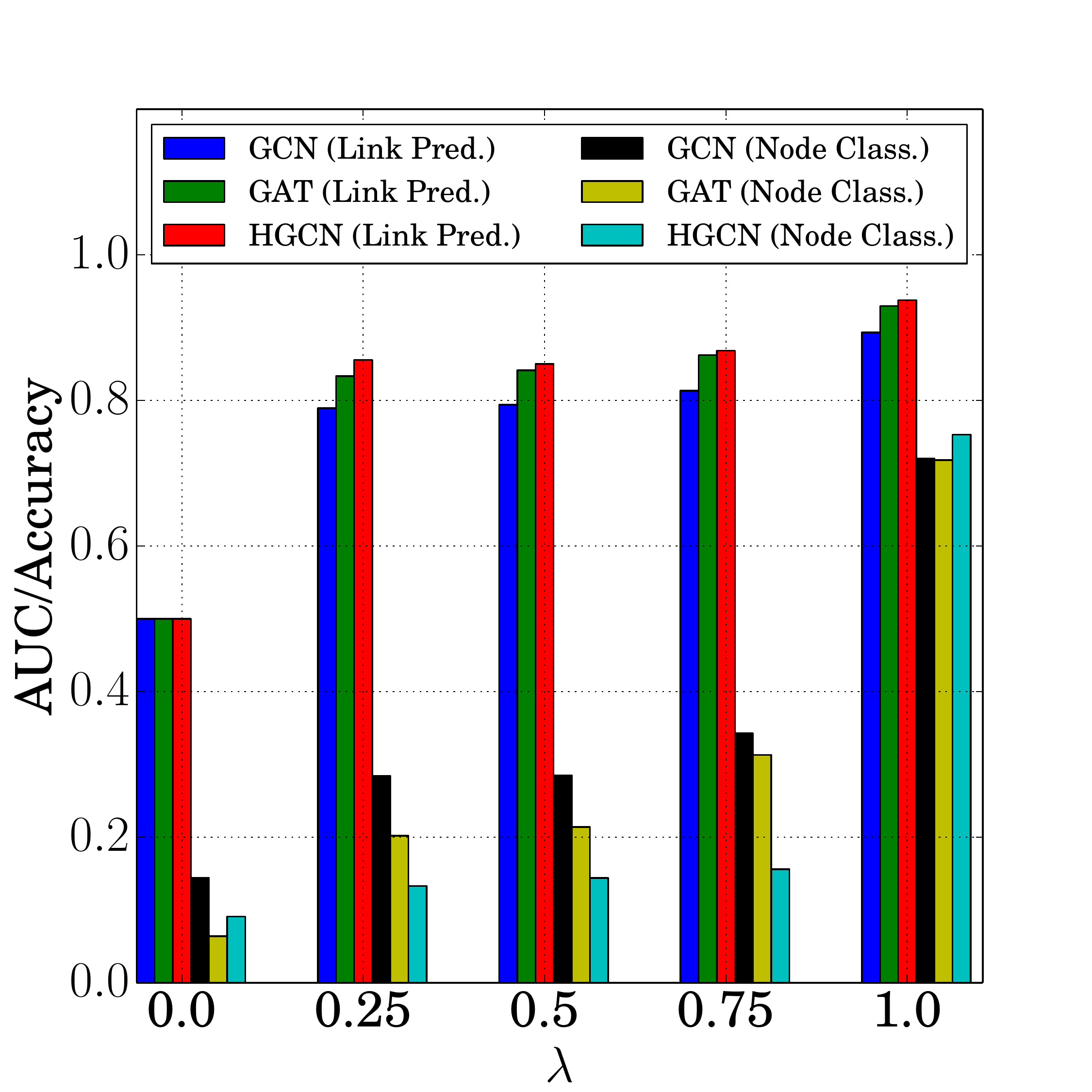} \label{fig1_topk}}
\subfigure[Citeseer]{\includegraphics[width=0.28\textwidth]{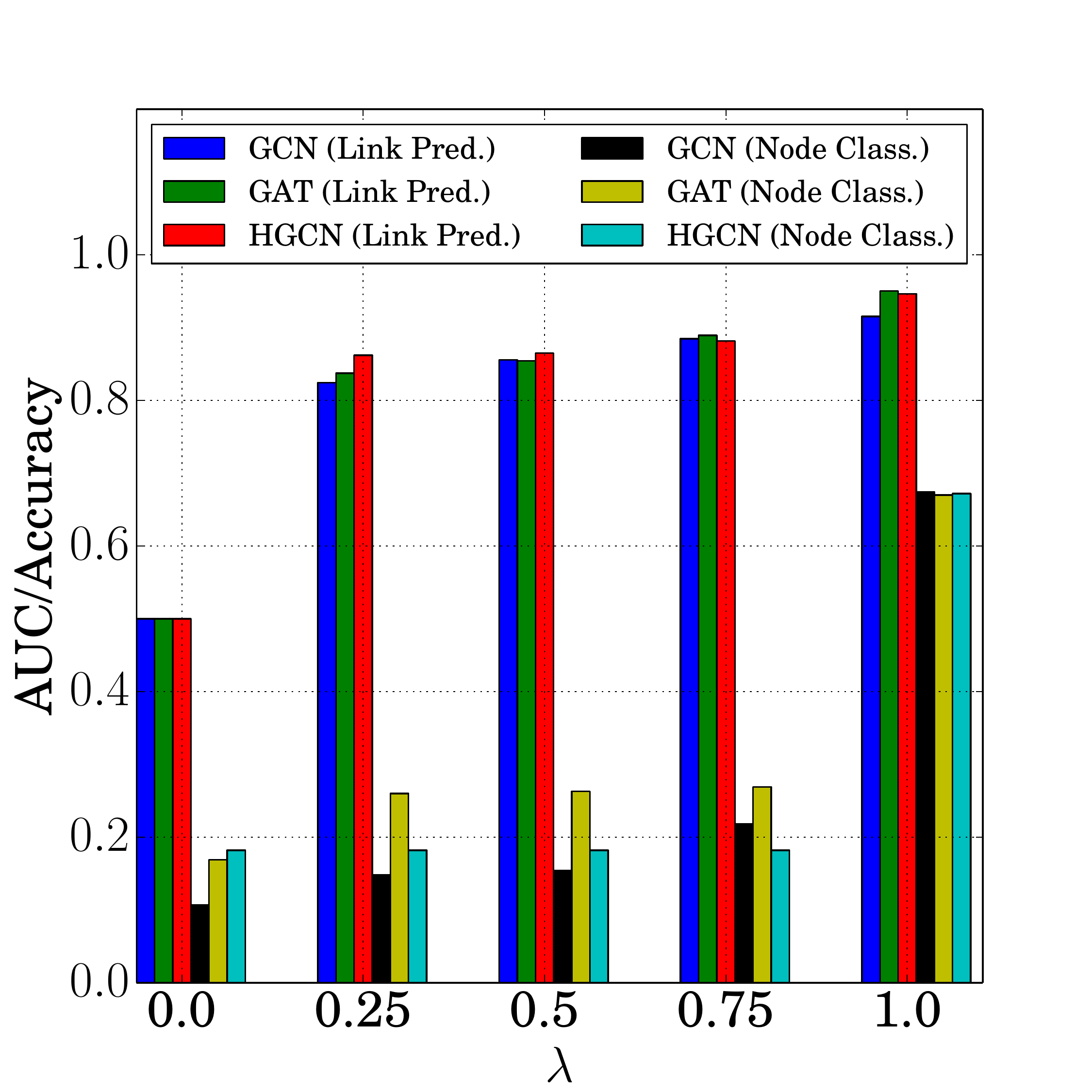} \label{fig2_topk}}
\subfigure[Pubmed]{\includegraphics[width=0.28\textwidth]{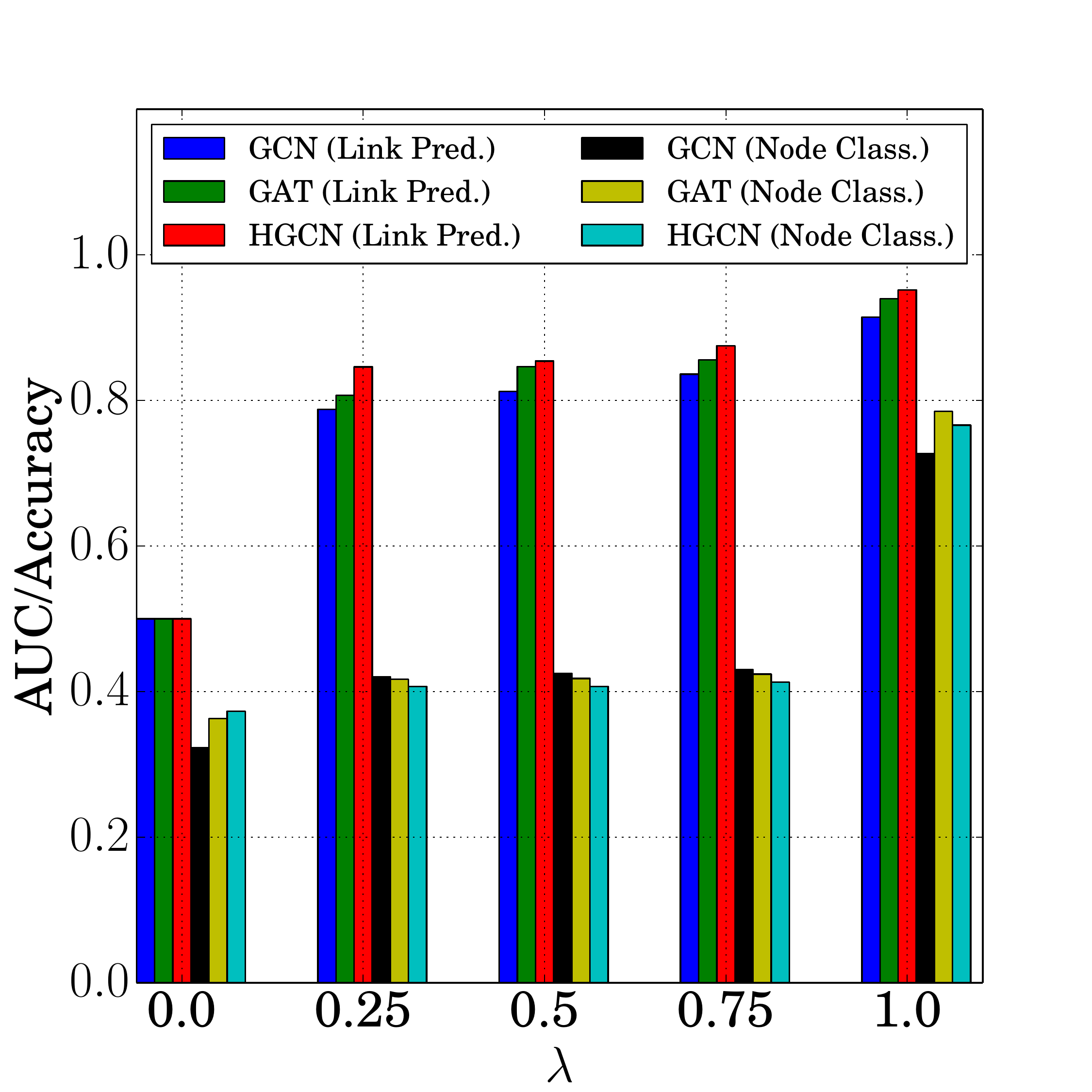} \label{fig2_topk}} 
\vspace{-3mm}
\caption{Impact of $\lambda$. Link prediction with node privacy protection.}
\label{fig:lambda_node_prot}
\vspace{-6mm}
\end{figure*}

\begin{figure*}[t]
\center
\subfigure[Cora]{\includegraphics[width=0.28\textwidth]{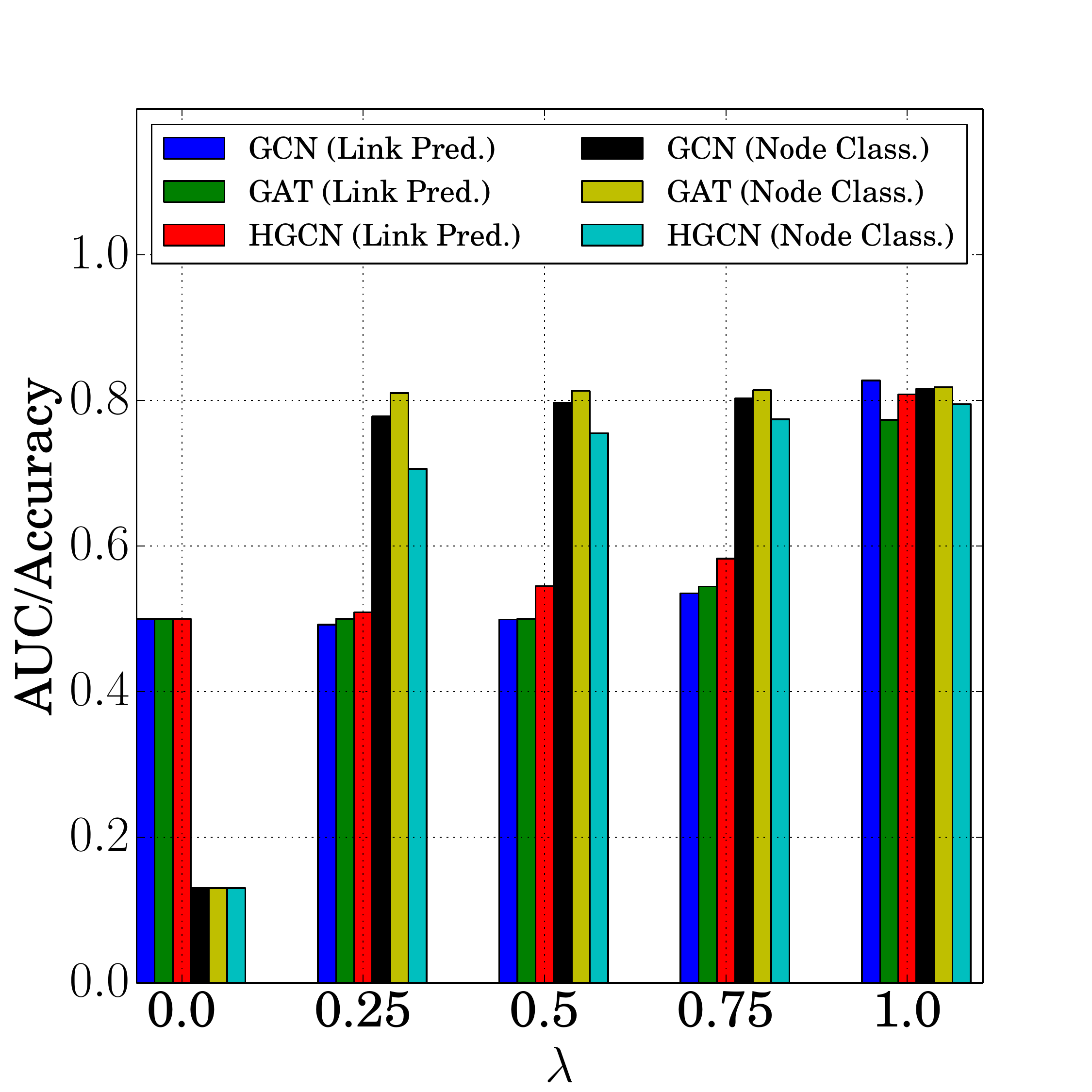} \label{fig1_topk}}
\subfigure[Citeseer]{\includegraphics[width=0.28\textwidth]{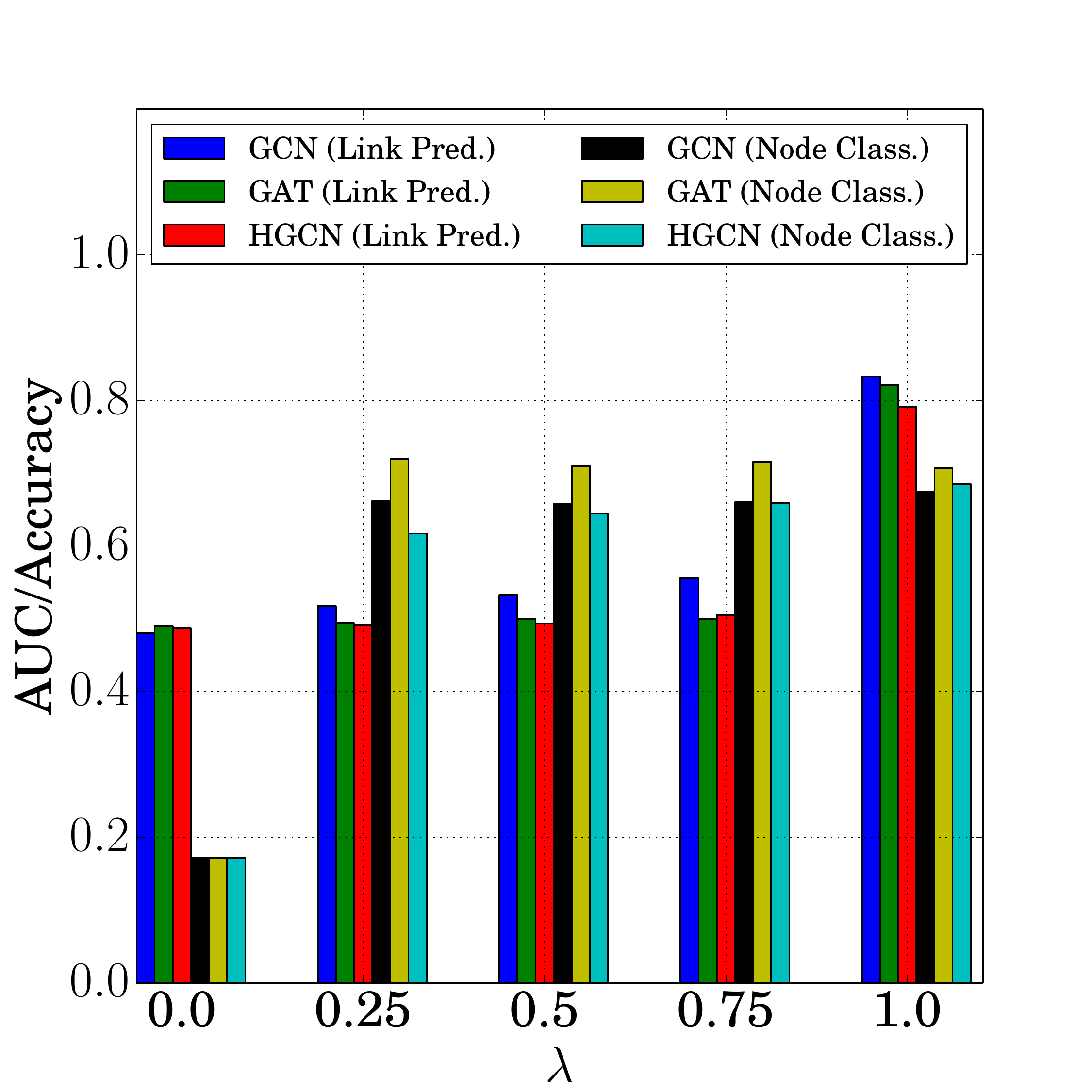} \label{fig2_topk}}
\subfigure[Pubmed]{\includegraphics[width=0.28\textwidth]{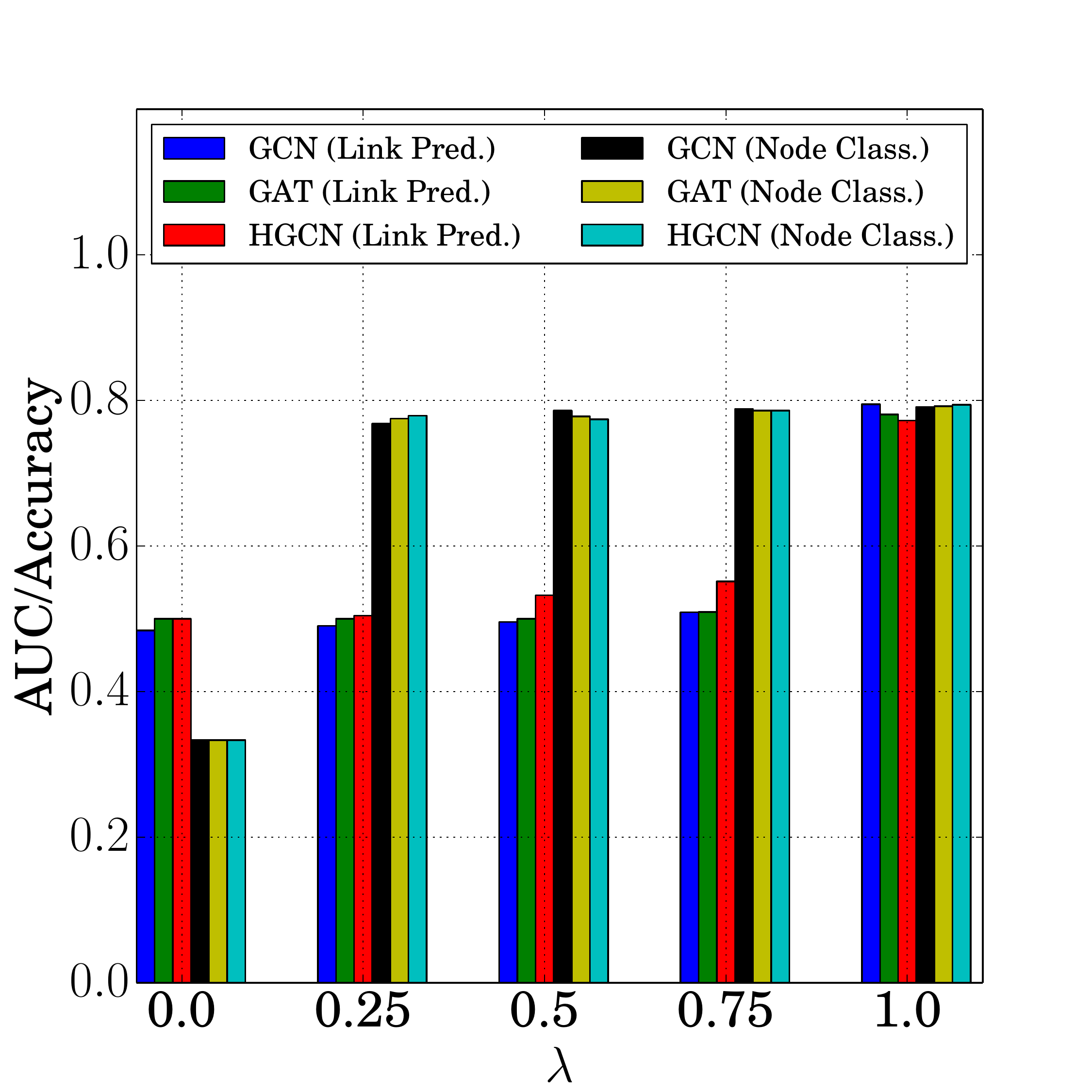} \label{fig2_topk}} 
\vspace{-3mm}
\caption{Impact of $\lambda$. Node classification with link privacy protection.}
\vspace{-2mm}
\label{fig:lambda_link_prot}
\end{figure*}

\subsection{Experimental Results}
\label{eval:results}

\subsubsection{Link prediction without/with node privacy protection}
In this experiment, we consider link prediction as the primary learning task, and node classification as the privacy protection task. 
Table~\ref{tbl:link_pred_wo_node_prot} shows the performance of the two tasks without node privacy protection by existing methods. Specifically, we use the three graph neural networks, i.e., GCN, GAT, and HGCN, to learn node representations, and use them to train a link predictor for link prediction. 
Next, we also leverage these node representations to train a node classifier. 
We have the following observations: 
1) All these methods achieve very high AUCs on the three graphs, i.e., almost all AUCs are above 90\%, demonstrating their effectiveness for link prediction.  
2) Although these node representations are not specially learnt for node classification, they can be used by the node classifier to accurately infer the node labels, thus leaking node privacy. For instance, all the methods obtain the accuracies around/above 70\% and they perform significantly better than random guessing.  

Table~\ref{tbl:link_pred_w_node_prot} shows the performance of the two tasks with node privacy protection.
Specifically, we use our framework to learn node representations, the link predictor, and the node classifier, simultaneously.  
We observe that our framework achieves an utility-privacy tradeoff. In particular, our framework has a tolerable AUC drop, compared with AUCs in Table~\ref{tbl:link_pred_w_node_prot}.  
However, our framework obtains much lower accuracies than those in  Table~\ref{tbl:link_pred_w_node_prot}. In some cases, the accuracies are close to random guessing, demonstrating a nearly perfect node privacy protection. 
The above results validate that our framework is effective for link prediction, as well as for protecting node privacy.

\subsubsection{Node classification with link privacy protection}
In this experiment, we consider node classification as the primary learning task, and link prediction as the privacy protection task. 
Table~\ref{tbl:node_pred_wo_link_prot} shows the performance of the two tasks without link privacy protection by existing methods. 
Similarly, we use the three graph neural networks to learn node representations, and use them to train a node classifier for node classification. 
We have similar observations as results shown in Table~\ref{tbl:link_pred_wo_node_prot}. 
First, all methods achieve promising accuracies on the three graphs, i.e., close to the results shown in~\cite{kipf2017semi,velivckovic2018graph,chami2019hyperbolic}.   
Next, we leverage these node representations to train a link predictor to infer link status. We observe that these methods obtain AUCs significantly larger than those obtained by random guessing, thus leaking link privacy seriously.  

Table~\ref{tbl:node_pred_w_link_prot} shows the performance of the two tasks with link privacy protection. Our framework achieves an utility-privacy tradeoff, similar to results in Table~\ref{tbl:node_pred_w_link_prot}. First, our framework has slightly accuracies degradation (around 1\%-4\%), compared with accuracies in Table~\ref{tbl:link_pred_w_node_prot}.  
Second, our framework obtains much lower AUCs and almost all of these AUCs are close to random guessing. 
The results again validate that our framework is effective for node classification, as well as for protecting link privacy.

\subsubsection{Impact of the trade-off factor $\lambda$}
\label{eval:impact_lambda}

In this experiment, we study the impact of the trade-off factor $\lambda$ in our framework.  
Figure~\ref{fig:lambda_node_prot}
and Figure~\ref{fig:lambda_link_prot} show the performance on the three graphs vs. different $\lambda$ for protecting node privacy and protecting link privacy, respectively. We have the following key observations: 
1) When $\lambda=0$, our framework only considers protecting node/link privacy, and achieves the lowest performance (i.e., close to random guessing) for inferring the node label/link status. 
However, the performance for the primary learning task is also the worst (i.e., random guessing). 
2) When $\lambda=1$, our framework only considers primary task learning and achieves the highest performance for the link prediction/node classification. However, it also obtains the highest performance for inferring the node label/link status, thus leaking the most information of nodes/link status. 
3) When $ 0 < \lambda < 1$, our framework considers both primary learning and privacy protection. 
We note that our framework is not sensitive to $\lambda$'s value in this range. That is, the performance of graph neural networks based on our framework for primary learning and privacy protection are relatively stable across all $\lambda$'s in this range.

\section{Conclusion}
\label{sec:conclu}

We propose the first framework for privacy-preserving representation learning on graphs from the mutual information perspective. 
Our framework includes a primary learning task and a privacy protection task. 
The goal is to learn node representations such that they can be used to achieve high performance for the primary learning task, while obtaining low performance for the privacy protection task (e.g., close to random guessing). 
We formally formulate our goal via mutual information objectives. However, mutual information is challenging to compute in practice. Motivated by mutual information neural estimation, we derive tractable variational bounds for the mutual information, and parameterize each bound via a neural network. 
Next, we train these neural networks to approximate the true mutual information and learn privacy-preserving node representations. 
We evaluate our framework on various 
graph datasets 
and show that our framework is effective for learning privacy-preserving node representations on graphs.

\noindent \textbf{Acknowledgements.}
We thank the anonymous reviewers for their constructive comments. This work is supported by the Amazon Research Award. Any opinions, findings and conclusions or
recommendations expressed in this material are those of the
author(s) and do not necessarily reflect the views of the funding
agencies.

\bibliographystyle{ACM-Reference-Format}
\bibliography{refs}


\begin{thebibliography}{39}


\ifx \showCODEN    \undefined \def \showCODEN     #1{\unskip}     \fi
\ifx \showDOI      \undefined \def \showDOI       #1{#1}\fi
\ifx \showISBNx    \undefined \def \showISBNx     #1{\unskip}     \fi
\ifx \showISBNxiii \undefined \def \showISBNxiii  #1{\unskip}     \fi
\ifx \showISSN     \undefined \def \showISSN      #1{\unskip}     \fi
\ifx \showLCCN     \undefined \def \showLCCN      #1{\unskip}     \fi
\ifx \shownote     \undefined \def \shownote      #1{#1}          \fi
\ifx \showarticletitle \undefined \def \showarticletitle #1{#1}   \fi
\ifx \showURL      \undefined \def \showURL       {\relax}        \fi
\providecommand\bibfield[2]{#2}
\providecommand\bibinfo[2]{#2}
\providecommand\natexlab[1]{#1}
\providecommand\showeprint[2][]{arXiv:#2}

\bibitem[\protect\citeauthoryear{Alemi, Fischer, Dillon, and Murphy}{Alemi
  et~al\mbox{.}}{2017}]%
        {alemi2017deep}
\bibfield{author}{\bibinfo{person}{Alexander~A Alemi}, \bibinfo{person}{Ian
  Fischer}, \bibinfo{person}{Joshua~V Dillon}, {and} \bibinfo{person}{Kevin
  Murphy}.} \bibinfo{year}{2017}\natexlab{}.
\newblock \showarticletitle{Deep variational information bottleneck}. In
  \bibinfo{booktitle}{\emph{ICLR}}.
\newblock


\bibitem[\protect\citeauthoryear{Belghazi, Baratin, Rajeshwar, Ozair, Bengio,
  Courville, and Hjelm}{Belghazi et~al\mbox{.}}{2018}]%
        {belghazi2018mutual}
\bibfield{author}{\bibinfo{person}{Mohamed~Ishmael Belghazi},
  \bibinfo{person}{Aristide Baratin}, \bibinfo{person}{Sai Rajeshwar},
  \bibinfo{person}{Sherjil Ozair}, \bibinfo{person}{Yoshua Bengio},
  \bibinfo{person}{Aaron Courville}, {and} \bibinfo{person}{Devon Hjelm}.}
  \bibinfo{year}{2018}\natexlab{}.
\newblock \showarticletitle{Mutual information neural estimation}. In
  \bibinfo{booktitle}{\emph{ICML}}.
\newblock


\bibitem[\protect\citeauthoryear{Cao, Lu, and Xu}{Cao et~al\mbox{.}}{2015}]%
        {cao2015grarep}
\bibfield{author}{\bibinfo{person}{Shaosheng Cao}, \bibinfo{person}{Wei Lu},
  {and} \bibinfo{person}{Qiongkai Xu}.} \bibinfo{year}{2015}\natexlab{}.
\newblock \showarticletitle{Grarep: Learning graph representations with global
  structural information}. In \bibinfo{booktitle}{\emph{CIKM}}.
\newblock


\bibitem[\protect\citeauthoryear{Chami, Ying, R{\'e}, and Leskovec}{Chami
  et~al\mbox{.}}{2019}]%
        {chami2019hyperbolic}
\bibfield{author}{\bibinfo{person}{Ines Chami}, \bibinfo{person}{Zhitao Ying},
  \bibinfo{person}{Christopher R{\'e}}, {and} \bibinfo{person}{Jure Leskovec}.}
  \bibinfo{year}{2019}\natexlab{}.
\newblock \showarticletitle{Hyperbolic graph convolutional neural networks}. In
  \bibinfo{booktitle}{\emph{NeurIPS}}.
\newblock


\bibitem[\protect\citeauthoryear{Chen, Duan, Houthooft, Schulman, Sutskever,
  and Abbeel}{Chen et~al\mbox{.}}{2016}]%
        {chen2016infogan}
\bibfield{author}{\bibinfo{person}{Xi Chen}, \bibinfo{person}{Yan Duan},
  \bibinfo{person}{Rein Houthooft}, \bibinfo{person}{John Schulman},
  \bibinfo{person}{Ilya Sutskever}, {and} \bibinfo{person}{Pieter Abbeel}.}
  \bibinfo{year}{2016}\natexlab{}.
\newblock \showarticletitle{Infogan: Interpretable representation learning by
  information maximizing generative adversarial nets}. In
  \bibinfo{booktitle}{\emph{NIPS}}.
\newblock


\bibitem[\protect\citeauthoryear{Cheng, Hao, Dai, Liu, Gan, and Carin}{Cheng
  et~al\mbox{.}}{2020}]%
        {cheng2020club}
\bibfield{author}{\bibinfo{person}{Pengyu Cheng}, \bibinfo{person}{Weituo Hao},
  \bibinfo{person}{Shuyang Dai}, \bibinfo{person}{Jiachang Liu},
  \bibinfo{person}{Zhe Gan}, {and} \bibinfo{person}{Lawrence Carin}.}
  \bibinfo{year}{2020}\natexlab{}.
\newblock \showarticletitle{CLUB: A Contrastive Log-ratio Upper Bound of Mutual
  Information}. In \bibinfo{booktitle}{\emph{ICML}}.
\newblock


\bibitem[\protect\citeauthoryear{Cui, Zhou, Yang, and Liu}{Cui
  et~al\mbox{.}}{2020}]%
        {cui2020adaptive}
\bibfield{author}{\bibinfo{person}{Ganqu Cui}, \bibinfo{person}{Jie Zhou},
  \bibinfo{person}{Cheng Yang}, {and} \bibinfo{person}{Zhiyuan Liu}.}
  \bibinfo{year}{2020}\natexlab{}.
\newblock \showarticletitle{Adaptive Graph Encoder for Attributed Graph
  Embedding}. In \bibinfo{booktitle}{\emph{KDD}}.
\newblock


\bibitem[\protect\citeauthoryear{Duran and Niepert}{Duran and Niepert}{2017}]%
        {duran2017learning}
\bibfield{author}{\bibinfo{person}{Alberto~Garcia Duran} {and}
  \bibinfo{person}{Mathias Niepert}.} \bibinfo{year}{2017}\natexlab{}.
\newblock \showarticletitle{Learning graph representations with embedding
  propagation}. In \bibinfo{booktitle}{\emph{NIPS}}.
\newblock


\bibitem[\protect\citeauthoryear{Duvenaud, Maclaurin, Iparraguirre, Bombarell,
  Hirzel, Aspuru-Guzik, and Adams}{Duvenaud et~al\mbox{.}}{2015}]%
        {duvenaud2015convolutional}
\bibfield{author}{\bibinfo{person}{David~K Duvenaud}, \bibinfo{person}{Dougal
  Maclaurin}, \bibinfo{person}{Jorge Iparraguirre}, \bibinfo{person}{Rafael
  Bombarell}, \bibinfo{person}{Timothy Hirzel}, \bibinfo{person}{Al{\'a}n
  Aspuru-Guzik}, {and} \bibinfo{person}{Ryan~P Adams}.}
  \bibinfo{year}{2015}\natexlab{}.
\newblock \showarticletitle{Convolutional networks on graphs for learning
  molecular fingerprints}. In \bibinfo{booktitle}{\emph{NIPS}}.
\newblock


\bibitem[\protect\citeauthoryear{Grover and Leskovec}{Grover and
  Leskovec}{2016}]%
        {grover2016node2vec}
\bibfield{author}{\bibinfo{person}{Aditya Grover} {and} \bibinfo{person}{Jure
  Leskovec}.} \bibinfo{year}{2016}\natexlab{}.
\newblock \showarticletitle{node2vec: Scalable feature learning for networks}.
  In \bibinfo{booktitle}{\emph{SIGKDD}}.
\newblock


\bibitem[\protect\citeauthoryear{Hamilton, Ying, and Leskovec}{Hamilton
  et~al\mbox{.}}{2017}]%
        {hamilton2017inductive}
\bibfield{author}{\bibinfo{person}{Will Hamilton}, \bibinfo{person}{Zhitao
  Ying}, {and} \bibinfo{person}{Jure Leskovec}.}
  \bibinfo{year}{2017}\natexlab{}.
\newblock \showarticletitle{Inductive representation learning on large graphs}.
  In \bibinfo{booktitle}{\emph{NIPS}}.
\newblock


\bibitem[\protect\citeauthoryear{Hjelm, Fedorov, Lavoie-Marchildon, Grewal,
  Bachman, Trischler, and Bengio}{Hjelm et~al\mbox{.}}{2019}]%
        {hjelm2019learning}
\bibfield{author}{\bibinfo{person}{R~Devon Hjelm}, \bibinfo{person}{Alex
  Fedorov}, \bibinfo{person}{Samuel Lavoie-Marchildon}, \bibinfo{person}{Karan
  Grewal}, \bibinfo{person}{Phil Bachman}, \bibinfo{person}{Adam Trischler},
  {and} \bibinfo{person}{Yoshua Bengio}.} \bibinfo{year}{2019}\natexlab{}.
\newblock \showarticletitle{Learning deep representations by mutual information
  estimation and maximization}. In \bibinfo{booktitle}{\emph{ICLR}}.
\newblock


\bibitem[\protect\citeauthoryear{Kipf and Welling}{Kipf and Welling}{2016}]%
        {kipf2016variational}
\bibfield{author}{\bibinfo{person}{Thomas~N Kipf} {and} \bibinfo{person}{Max
  Welling}.} \bibinfo{year}{2016}\natexlab{}.
\newblock \showarticletitle{Variational graph auto-encoders}. In
  \bibinfo{booktitle}{\emph{NIPS Workshop}}.
\newblock


\bibitem[\protect\citeauthoryear{Kipf and Welling}{Kipf and Welling}{2017}]%
        {kipf2017semi}
\bibfield{author}{\bibinfo{person}{Thomas~N Kipf} {and} \bibinfo{person}{Max
  Welling}.} \bibinfo{year}{2017}\natexlab{}.
\newblock \showarticletitle{Semi-supervised classification with graph
  convolutional networks}.
\newblock \bibinfo{journal}{\emph{ICLR}} (\bibinfo{year}{2017}).
\newblock


\bibitem[\protect\citeauthoryear{Krioukov, Papadopoulos, Kitsak, Vahdat, and
  Bogun{\'a}}{Krioukov et~al\mbox{.}}{2010}]%
        {krioukov2010hyperbolic}
\bibfield{author}{\bibinfo{person}{Dmitri Krioukov},
  \bibinfo{person}{Fragkiskos Papadopoulos}, \bibinfo{person}{Maksim Kitsak},
  \bibinfo{person}{Amin Vahdat}, {and} \bibinfo{person}{Mari{\'a}n
  Bogun{\'a}}.} \bibinfo{year}{2010}\natexlab{}.
\newblock \showarticletitle{Hyperbolic geometry of complex networks}.
\newblock \bibinfo{journal}{\emph{Physical Review E}} (\bibinfo{year}{2010}).
\newblock


\bibitem[\protect\citeauthoryear{Li, Duan, Yang, Chen, and Yang}{Li
  et~al\mbox{.}}{2020}]%
        {li2020tiprdc}
\bibfield{author}{\bibinfo{person}{Ang Li}, \bibinfo{person}{Yixiao Duan},
  \bibinfo{person}{Huanrui Yang}, \bibinfo{person}{Yiran Chen}, {and}
  \bibinfo{person}{Jianlei Yang}.} \bibinfo{year}{2020}\natexlab{}.
\newblock \showarticletitle{TIPRDC: task-independent privacy-respecting data
  crowdsourcing framework for deep learning with anonymized intermediate
  representations}. In \bibinfo{booktitle}{\emph{KDD}}.
\newblock


\bibitem[\protect\citeauthoryear{Liu, Nickel, and Kiela}{Liu
  et~al\mbox{.}}{2019}]%
        {liu2019hyperbolic}
\bibfield{author}{\bibinfo{person}{Qi Liu}, \bibinfo{person}{Maximilian
  Nickel}, {and} \bibinfo{person}{Douwe Kiela}.}
  \bibinfo{year}{2019}\natexlab{}.
\newblock \showarticletitle{Hyperbolic graph neural networks}. In
  \bibinfo{booktitle}{\emph{NeurIPS}}.
\newblock


\bibitem[\protect\citeauthoryear{Ma, Wang, Aggarwal, and Tang}{Ma
  et~al\mbox{.}}{2019}]%
        {ma2019graph}
\bibfield{author}{\bibinfo{person}{Yao Ma}, \bibinfo{person}{Suhang Wang},
  \bibinfo{person}{Charu~C Aggarwal}, {and} \bibinfo{person}{Jiliang Tang}.}
  \bibinfo{year}{2019}\natexlab{}.
\newblock \showarticletitle{Graph convolutional networks with eigenpooling}. In
  \bibinfo{booktitle}{\emph{KDD}}.
\newblock


\bibitem[\protect\citeauthoryear{Nickel and Kiela}{Nickel and Kiela}{2017}]%
        {nickel2017poincare}
\bibfield{author}{\bibinfo{person}{Maximillian Nickel} {and}
  \bibinfo{person}{Douwe Kiela}.} \bibinfo{year}{2017}\natexlab{}.
\newblock \showarticletitle{Poincar{\'e} embeddings for learning hierarchical
  representations}. In \bibinfo{booktitle}{\emph{NIPS}}.
\newblock


\bibitem[\protect\citeauthoryear{Oord, Li, and Vinyals}{Oord
  et~al\mbox{.}}{2018}]%
        {oord2018representation}
\bibfield{author}{\bibinfo{person}{Aaron van~den Oord}, \bibinfo{person}{Yazhe
  Li}, {and} \bibinfo{person}{Oriol Vinyals}.} \bibinfo{year}{2018}\natexlab{}.
\newblock \showarticletitle{Representation learning with contrastive predictive
  coding}.
\newblock \bibinfo{journal}{\emph{arXiv}} (\bibinfo{year}{2018}).
\newblock


\bibitem[\protect\citeauthoryear{Peng, Huang, Luo, Zheng, Rong, Xu, and
  Huang}{Peng et~al\mbox{.}}{2020}]%
        {peng2020graph}
\bibfield{author}{\bibinfo{person}{Zhen Peng}, \bibinfo{person}{Wenbing Huang},
  \bibinfo{person}{Minnan Luo}, \bibinfo{person}{Qinghua Zheng},
  \bibinfo{person}{Yu Rong}, \bibinfo{person}{Tingyang Xu}, {and}
  \bibinfo{person}{Junzhou Huang}.} \bibinfo{year}{2020}\natexlab{}.
\newblock \showarticletitle{Graph Representation Learning via Graphical Mutual
  Information Maximization}. In \bibinfo{booktitle}{\emph{WWW}}.
\newblock


\bibitem[\protect\citeauthoryear{Perozzi, Al-Rfou, and Skiena}{Perozzi
  et~al\mbox{.}}{2014}]%
        {perozzi2014deepwalk}
\bibfield{author}{\bibinfo{person}{Bryan Perozzi}, \bibinfo{person}{Rami
  Al-Rfou}, {and} \bibinfo{person}{Steven Skiena}.}
  \bibinfo{year}{2014}\natexlab{}.
\newblock \showarticletitle{Deepwalk: Online learning of social
  representations}. In \bibinfo{booktitle}{\emph{SIGKDD}}.
\newblock


\bibitem[\protect\citeauthoryear{Poole, Ozair, Oord, Alemi, and Tucker}{Poole
  et~al\mbox{.}}{2019}]%
        {poole2019variational}
\bibfield{author}{\bibinfo{person}{Ben Poole}, \bibinfo{person}{Sherjil Ozair},
  \bibinfo{person}{Aaron van~den Oord}, \bibinfo{person}{Alexander~A Alemi},
  {and} \bibinfo{person}{George Tucker}.} \bibinfo{year}{2019}\natexlab{}.
\newblock \showarticletitle{On variational bounds of mutual information}. In
  \bibinfo{booktitle}{\emph{ICML}}.
\newblock


\bibitem[\protect\citeauthoryear{Qiu, Dong, Ma, Li, Wang, and Tang}{Qiu
  et~al\mbox{.}}{2018}]%
        {qiu2018network}
\bibfield{author}{\bibinfo{person}{Jiezhong Qiu}, \bibinfo{person}{Yuxiao
  Dong}, \bibinfo{person}{Hao Ma}, \bibinfo{person}{Jian Li},
  \bibinfo{person}{Kuansan Wang}, {and} \bibinfo{person}{Jie Tang}.}
  \bibinfo{year}{2018}\natexlab{}.
\newblock \showarticletitle{Network embedding as matrix factorization: Unifying
  deepwalk, line, pte, and node2vec}. In \bibinfo{booktitle}{\emph{WSDM}}.
\newblock


\bibitem[\protect\citeauthoryear{Qu, Bengio, and Tang}{Qu
  et~al\mbox{.}}{2019}]%
        {qu2019gmnn}
\bibfield{author}{\bibinfo{person}{Meng Qu}, \bibinfo{person}{Yoshua Bengio},
  {and} \bibinfo{person}{Jian Tang}.} \bibinfo{year}{2019}\natexlab{}.
\newblock \showarticletitle{GMNN: Graph Markov Neural Networks}. In
  \bibinfo{booktitle}{\emph{ICML}}.
\newblock


\bibitem[\protect\citeauthoryear{Ribeiro, Saverese, and Figueiredo}{Ribeiro
  et~al\mbox{.}}{2017}]%
        {ribeiro2017struc2vec}
\bibfield{author}{\bibinfo{person}{Leonardo~FR Ribeiro},
  \bibinfo{person}{Pedro~HP Saverese}, {and} \bibinfo{person}{Daniel~R
  Figueiredo}.} \bibinfo{year}{2017}\natexlab{}.
\newblock \showarticletitle{struc2vec: Learning node representations from
  structural identity}. In \bibinfo{booktitle}{\emph{KDD}}.
\newblock


\bibitem[\protect\citeauthoryear{Rossi, Zhou, and Ahmed}{Rossi
  et~al\mbox{.}}{2018}]%
        {rossi2018deep}
\bibfield{author}{\bibinfo{person}{Ryan~A Rossi}, \bibinfo{person}{Rong Zhou},
  {and} \bibinfo{person}{Nesreen~K Ahmed}.} \bibinfo{year}{2018}\natexlab{}.
\newblock \showarticletitle{Deep inductive network representation learning}. In
  \bibinfo{booktitle}{\emph{WWW}}.
\newblock


\bibitem[\protect\citeauthoryear{Sen, Namata, Bilgic, and et~al.}{Sen
  et~al\mbox{.}}{2008}]%
        {sen2008collective}
\bibfield{author}{\bibinfo{person}{Prithviraj Sen}, \bibinfo{person}{Galileo
  Namata}, \bibinfo{person}{Mustafa Bilgic}, {and} \bibinfo{person}{et al.}}
  \bibinfo{year}{2008}\natexlab{}.
\newblock \showarticletitle{Collective classification in network data}.
\newblock \bibinfo{journal}{\emph{AI magazine}} (\bibinfo{year}{2008}).
\newblock


\bibitem[\protect\citeauthoryear{Sun, Hoffmann, Verma, and Tang}{Sun
  et~al\mbox{.}}{2020}]%
        {sun2020infograph}
\bibfield{author}{\bibinfo{person}{Fan-Yun Sun}, \bibinfo{person}{Jordan
  Hoffmann}, \bibinfo{person}{Vikas Verma}, {and} \bibinfo{person}{Jian Tang}.}
  \bibinfo{year}{2020}\natexlab{}.
\newblock \showarticletitle{Infograph: Unsupervised and semi-supervised
  graph-level representation learning via mutual information maximization}. In
  \bibinfo{booktitle}{\emph{ICLR}}.
\newblock


\bibitem[\protect\citeauthoryear{Tang, Qu, Wang, Zhang, Yan, and Mei}{Tang
  et~al\mbox{.}}{2015}]%
        {tang2015line}
\bibfield{author}{\bibinfo{person}{Jian Tang}, \bibinfo{person}{Meng Qu},
  \bibinfo{person}{Mingzhe Wang}, \bibinfo{person}{Ming Zhang},
  \bibinfo{person}{Jun Yan}, {and} \bibinfo{person}{Qiaozhu Mei}.}
  \bibinfo{year}{2015}\natexlab{}.
\newblock \showarticletitle{Line: Large-scale information network embedding}.
  In \bibinfo{booktitle}{\emph{WWW}}.
\newblock


\bibitem[\protect\citeauthoryear{Tu, Zhang, Liu, and Sun}{Tu
  et~al\mbox{.}}{2016}]%
        {tu2016max}
\bibfield{author}{\bibinfo{person}{Cunchao Tu}, \bibinfo{person}{Weicheng
  Zhang}, \bibinfo{person}{Zhiyuan Liu}, {and} \bibinfo{person}{Maosong Sun}.}
  \bibinfo{year}{2016}\natexlab{}.
\newblock \showarticletitle{Max-margin DeepWalk: discriminative learning of
  network representation}. In \bibinfo{booktitle}{\emph{IJCAI}}.
\newblock


\bibitem[\protect\citeauthoryear{Veli{\v{c}}kovi{\'c}, Cucurull, Casanova,
  Romero, Lio, and Bengio}{Veli{\v{c}}kovi{\'c} et~al\mbox{.}}{2018}]%
        {velivckovic2018graph}
\bibfield{author}{\bibinfo{person}{Petar Veli{\v{c}}kovi{\'c}},
  \bibinfo{person}{Guillem Cucurull}, \bibinfo{person}{Arantxa Casanova},
  \bibinfo{person}{Adriana Romero}, \bibinfo{person}{Pietro Lio}, {and}
  \bibinfo{person}{Yoshua Bengio}.} \bibinfo{year}{2018}\natexlab{}.
\newblock \showarticletitle{Graph attention networks}. In
  \bibinfo{booktitle}{\emph{ICLR}}.
\newblock


\bibitem[\protect\citeauthoryear{Velickovic, Fedus, Hamilton, Li{\`o}, Bengio,
  and Hjelm}{Velickovic et~al\mbox{.}}{2019}]%
        {velickovic2019deep}
\bibfield{author}{\bibinfo{person}{Petar Velickovic}, \bibinfo{person}{William
  Fedus}, \bibinfo{person}{William~L Hamilton}, \bibinfo{person}{Pietro
  Li{\`o}}, \bibinfo{person}{Yoshua Bengio}, {and} \bibinfo{person}{R~Devon
  Hjelm}.} \bibinfo{year}{2019}\natexlab{}.
\newblock \showarticletitle{Deep Graph Infomax.}. In
  \bibinfo{booktitle}{\emph{ICLR}}.
\newblock


\bibitem[\protect\citeauthoryear{Wu, Zhang, Souza~Jr, Fifty, Yu, and
  Weinberger}{Wu et~al\mbox{.}}{2019b}]%
        {wu2019simplifying}
\bibfield{author}{\bibinfo{person}{Felix Wu}, \bibinfo{person}{Tianyi Zhang},
  \bibinfo{person}{Amauri Holanda~de Souza~Jr}, \bibinfo{person}{Christopher
  Fifty}, \bibinfo{person}{Tao Yu}, {and} \bibinfo{person}{Kilian~Q
  Weinberger}.} \bibinfo{year}{2019}\natexlab{b}.
\newblock \showarticletitle{Simplifying graph convolutional networks}. In
  \bibinfo{booktitle}{\emph{ICML}}.
\newblock


\bibitem[\protect\citeauthoryear{Wu, He, and Xu}{Wu et~al\mbox{.}}{2019a}]%
        {wu2019net}
\bibfield{author}{\bibinfo{person}{Jun Wu}, \bibinfo{person}{Jingrui He}, {and}
  \bibinfo{person}{Jiejun Xu}.} \bibinfo{year}{2019}\natexlab{a}.
\newblock \showarticletitle{Demo-Net: Degree-specific graph neural networks for
  node and graph classification}. In \bibinfo{booktitle}{\emph{KDD}}.
\newblock


\bibitem[\protect\citeauthoryear{Xu, Hu, Leskovec, and Jegelka}{Xu
  et~al\mbox{.}}{2019}]%
        {xu2018powerful}
\bibfield{author}{\bibinfo{person}{Keyulu Xu}, \bibinfo{person}{Weihua Hu},
  \bibinfo{person}{Jure Leskovec}, {and} \bibinfo{person}{Stefanie Jegelka}.}
  \bibinfo{year}{2019}\natexlab{}.
\newblock \showarticletitle{How powerful are graph neural networks?}. In
  \bibinfo{booktitle}{\emph{ICLR}}.
\newblock


\bibitem[\protect\citeauthoryear{Xu, Li, Tian, Sonobe, Kawarabayashi, and
  Jegelka}{Xu et~al\mbox{.}}{2018}]%
        {xu2018representation}
\bibfield{author}{\bibinfo{person}{Keyulu Xu}, \bibinfo{person}{Chengtao Li},
  \bibinfo{person}{Yonglong Tian}, \bibinfo{person}{Tomohiro Sonobe},
  \bibinfo{person}{Ken-ichi Kawarabayashi}, {and} \bibinfo{person}{Stefanie
  Jegelka}.} \bibinfo{year}{2018}\natexlab{}.
\newblock \showarticletitle{Representation learning on graphs with jumping
  knowledge networks}. In \bibinfo{booktitle}{\emph{ICML}}.
\newblock


\bibitem[\protect\citeauthoryear{Yang, Cohen, and Salakhudinov}{Yang
  et~al\mbox{.}}{2016}]%
        {yang2016revisiting}
\bibfield{author}{\bibinfo{person}{Zhilin Yang}, \bibinfo{person}{William
  Cohen}, {and} \bibinfo{person}{Ruslan Salakhudinov}.}
  \bibinfo{year}{2016}\natexlab{}.
\newblock \showarticletitle{Revisiting semi-supervised learning with graph
  embeddings}. In \bibinfo{booktitle}{\emph{ICML}}.
\newblock


\bibitem[\protect\citeauthoryear{Zhang and Chen}{Zhang and Chen}{2018}]%
        {zhang2018link}
\bibfield{author}{\bibinfo{person}{Muhan Zhang} {and} \bibinfo{person}{Yixin
  Chen}.} \bibinfo{year}{2018}\natexlab{}.
\newblock \showarticletitle{Link prediction based on graph neural networks}. In
  \bibinfo{booktitle}{\emph{NIPS}}.
\newblock


\end{thebibliography}

\appendix
\section{Algorithmic Pseudo Code}

\begin{algorithm}[h]
\caption{Link prediction with node privacy protection}
\begin{flushleft}
{\bf Input:}  Graph $G$, $\mathcal{V}_L = \{ \mathbf{x}_u, y_u\}$,  $\mathcal{E}_p = \{\mathbf{x}_u, \mathbf{x}_v, A_{uv}=1 \}$, 
$\mathcal{E}_n = \{\mathbf{x}_u, \mathbf{x}_v, A_{uv}=0 \}$, 
trade-off factor $\lambda$, 
\#gradient steps $I$,
\#rounds $T$. \\
{\bf Output:} Network parameters: $\theta^*, \phi^*, \psi^*$.
\end{flushleft}

\begin{algorithmic}[1]
\State Initialize $\theta, \phi, \psi$ for the embedding network $f_\theta$, 
link predictor $h_\phi$, and node classifier $g_\psi$;
\State Initialize learning rates $lr_1, lr_2, lr_3$;
\State Initialize $t \leftarrow 0$;

\For{$t < T$}
    \State $L_1 = \sum_{(u, v) \in \mathcal{E}_p \cup \mathcal{E}_n} CE(h_\phi(f_\theta(\mathbf{x}_u), f_\theta(\mathbf{x}_v)), A_{uv})$;
    \State $L_2 = \sum_{v \in \mathcal{V}_L} CE(g_\psi(f_\theta(\mathbf{x}_v)), y_v)$;
    
    \State Initialize $i \leftarrow 0$;
    \For{$i < I$}
        \State $\phi \leftarrow \phi - lr_1 \cdot \frac{\partial L_1}{\partial \phi}$;
        \State $\psi \leftarrow \psi + lr_2 \cdot \frac{\partial L_2}{\partial \psi}$;
        \State $\theta \leftarrow \theta - lr_3 \cdot \frac{\partial (\lambda L_1 - (1-\lambda) L2)}{\partial \theta} $;
        \State $i \leftarrow i+1$;
    \EndFor
    \State $t \leftarrow t+1$
\EndFor
\State \Return $\theta, \phi, \psi$.
\end{algorithmic}
\label{alg:P1}
\end{algorithm}

\begin{algorithm}[h]
\caption{Node classification with link privacy protection}
\begin{flushleft}
{\bf Input:}  Graph $G$, $\mathcal{V}_L = \{ \mathbf{x}_u, y_u\}$,  $\mathcal{E}_p = \{\mathbf{x}_u, \mathbf{x}_v, A_{uv}=1 \}$, 
$\mathcal{E}_n = \{\mathbf{x}_u, \mathbf{x}_v, A_{uv}=0 \}$, 
trade-off factor $\lambda$, 
\#gradient steps $I$,
\#rounds $T$. \\
{\bf Output:} Network parameters: $\theta^*, \phi^*, \psi^*$.
\end{flushleft}

\begin{algorithmic}[1]
\State Initialize $\theta, \phi, \psi$ for the embedding network $f_\theta$, 
link predictor $h_\phi$, and node classifier $g_\psi$;

\State Initialize learning rates $lr_1, lr_2, lr_3$;

\State Initialize $t \leftarrow 0$.

\For{$t < T$}

    \State $L_1 = \sum_{v \in \mathcal{V}_L} CE(g_\psi(f_\theta(\mathbf{x}_v)), y_v)$;
    
    \State $L_2 = \sum_{(u, v) \in \mathcal{E}_p \cup \mathcal{E}_n} CE(h_\phi(f_\theta(\mathbf{x}_u), f_\theta(\mathbf{x}_v)), A_{uv})$; 
    
    \State Initialize $i \leftarrow 0$;
    
    \For{$i < I$}
    
    \State $\psi \leftarrow \psi - lr_1 \cdot \frac{\partial L_1}{\partial \psi}$;
    
    \State $\phi \leftarrow \phi + lr_2 \cdot \frac{\partial L_2}{\partial \phi}$;

    \State $\theta \leftarrow \theta - lr_3 \cdot \frac{\partial (\lambda L_1- (1-\lambda) L2)}{\partial \theta} $;
    
    \State $i \leftarrow i+1$;
    \EndFor
    
    \State $t \leftarrow t+1$
\EndFor

\State \Return $\theta, \phi, \psi$.
\end{algorithmic}
\label{alg:P2}
\end{algorithm}

\end{document}